\documentclass{article}
\usepackage{iclr2025_conference,times}
\iclrfinalcopy
\usepackage{amsmath,amssymb,amsfonts, bm} 
\usepackage{algorithm}
\usepackage[noend]{algorithmic}
\usepackage{caption}
\usepackage{bbold}
\usepackage{subcaption}
\usepackage{nicematrix}
\usepackage{booktabs}
\usepackage{multirow}
\usepackage{tikz}
\usepackage{microtype}
\usepackage{hyperref}  
\usepackage{cleveref}
\usepackage{url}   
\usepackage{natbib}
\usepackage{float} 
\usepackage{etoc}
\usepackage{enumitem}
\usepackage{pifont}
\usepackage{tcolorbox}
\usepackage{tabularx}

\newcommand{\NN}{\mathbb{N}}

\newcommand{\mb}{\mathsf{B}}
\newcommand{\N}{\mathbb{N}}

\newcommand{\J}{\mathcal{J}}

\newtheorem{remark}{Remark}

\newcommand{\R}{\mathbb{R}}
\newcommand{\optarraystretch}{1}
\newcommand{\sgd}{\text{SGD}}
\newcommand{\momentum}{\text{Momentum}}
\newcommand{\nesterov}{\text{Nesterov}}
\newcommand{\adam}{\text{Adam}}
\newcommand{\nadam}{\text{NAdam}}
\newcommand{\rmspropmom}{\text{RMSprop + Momentum}}
\newcommand{\rmsprop}{\text{RMSprop}}
\newcommand{\inna}{\text{INNA}}
\newcommand{\dinadam}{\text{DINAdam}}
\newcommand{\INNAprop}{\text{INNAprop}}

\newcommand{\sgdheader}{$\sgd(\gamma_k)$}
\newcommand{\sgdbody}{
$
\begin{aligned}
    \theta_{k+1} = \theta_{k} - \gamma_k g_k
\end{aligned}
$
}
\newcommand{\momheader}{$\momentum(\gamma_k, \beta_1)$}
\newcommand{\mombody}{
$
\begin{aligned}
    &v_0 = 0  \\
    &v_{k+1} = \beta_1 v_{k} + (1 - \beta_1) g_k\\
    &\theta_{k+1} = \theta_{k} - \gamma_k v_{k+1}
\end{aligned}
$
}

\newcommand{\rmspropmomheader}{$\rmspropmom(\gamma_k, \beta_1, \beta_2, \epsilon)$}
\newcommand{\rmspropmombody}{
$
\begin{aligned}
    &v_0 = 1 \text{,} \, m_0 = 0 \\
    &v_{k+1} = \beta_2 v_{k} + (1 - \beta_2) g_k^2 \\
    &m_{k+1} = \beta_1 m_{k} + \frac{g_k}{\sqrt{v_{k+1} + \epsilon}}\\
    &\theta_{k+1} = \theta_{k} - \gamma_k m_{k+1}
\end{aligned}
$
}

\newcommand{\adamheader}{$\adam(\gamma_k, \beta_1, \beta_2, \epsilon)$}
\newcommand{\adambody}{
$
\begin{aligned}
    &m_0 = 0 \text{,} \, v_0 = 0 \\
    &m_{k+1} = \beta_1 m_{k} + (1 - \beta_1) g_k &\\
    &v_{k+1} = \beta_2 v_{k} + (1 - \beta_2) g_k^2\\
    &\theta_{k+1} = \theta_{k} - \gamma_k \frac{m_{k+1}}{\sqrt{v_{k+1}} + \epsilon}
\end{aligned}
$
}
\newcommand{\nadamheader}{$\nadam(\gamma_k, \psi, \beta_1, \beta_2, \epsilon)$}
\newcommand{\nadambody}{
$
\begin{aligned}
    &m_0 = 0 \text{,} \, v_0 = 0 \\
    &\mu_k = \beta_1(1 - \frac{1}{2}0.96^{k \psi}) \\
    &m_{k+1} = \beta_1 m_{k} + (1 - \beta_1) g_k \\
    &v_{k+1} = \beta_2 v_{k} + (1 - \beta_2) g_k^2\\
    &\theta_{k+1} = \theta_{k} - \gamma_k \frac{\mu_{k+1} m_{k+1} + (1 - \mu_k) g_k}{\sqrt{v_{k+1}} + \epsilon}
\end{aligned}
$
}

\newcommand{\innaheader}{$\inna(\gamma_k, \alpha, \beta)$}
\newcommand{\innabody}{
$
\begin{aligned}
    &\psi_0 = (1 - \alpha \beta) \theta_0\\
    &\psi_{k+1} = \psi_k + \gamma_k\left(( \frac{1}{\beta} - \alpha) \theta_k  - \frac{1}{\beta} \psi_k \right) \\
    &\theta_{k+1} = \theta_k + \gamma_k \left(( \frac{1}{\beta} - \alpha) \theta_k - \frac{1}{\beta} \psi_k - \beta g_k \right)
\end{aligned}
$
}

\renewcommand{\algorithmiccomment}[1]{\bgroup\hfill $\triangleright$ ~#1\egroup}

\etocdepthtag.toc{mtsection}
\etocsettagdepth{mtsection}{subsection}
\etocsettagdepth{mtappendix}{none}
\title{A second-order-like optimizer with adaptive gradient scaling for deep learning}
\author{Jérôme Bolte\textsuperscript{1,2}, Ryan Boustany\textsuperscript{1,2,3},  Edouard Pauwels \textsuperscript{1,2} \& Andrei Purica\textsuperscript{3} \\
\textsuperscript{1} Toulouse School of Economics \\
\textsuperscript{2} Université de Toulouse \\
\textsuperscript{3} Thales LAS France \\
\texttt{\{{jerome.bolte, ryan.boustany, edouard.pauwels\}@ut-capitole.fr}}, \\ \texttt{andrei.purica@thalesgroup.com}
}

\begin{document}

\maketitle

\begin{abstract}
In this empirical article, we introduce INNAprop, an optimization algorithm that combines the INNA method with the RMSprop adaptive gradient scaling. It leverages second-order information and rescaling while keeping the memory requirements of standard DL methods as AdamW or SGD with momentum. After giving geometrical insights, we evaluate INNAprop on CIFAR-10, Food101, and ImageNet with ResNets, VGG, DenseNet, and ViT, and on GPT-2 (OpenWebText) train from scratch and with LoRA fine-tuning (E2E). INNAprop consistently matches or outperforms AdamW both in training speed and accuracy, with minimal hyperparameter tuning in large-scale settings. Our code is publicly available at \url{https://github.com/innaprop/innaprop}.
\end{abstract}
\section{Introduction}
\label{sec:intro}
As deep learning models grow in size, massive computational resources are needed for training, representing significant challenges in terms of financial costs, energy consumption, and processing time \citep{susnjak2024over,varoquaux2024hype}. According to the UN’s Environment Programme Training, the Big Tech sector produced between two and three percent of the world’s carbon emissions in 2021; some estimations for the year 2023 go beyond 4\%, see the latest Stand.earth reports, and also \citep{schwartz2020green, strubell2020energy, patterson2021carbon} for related issues. For instance, training GPT-3 is estimated to require 1,287 megawatt-hours (MWh) of electricity, equivalent to the annual usage of over 100 U.S. households \citep{anthony2020carbontracker, patterson2021carbon}. Similarly, the financial cost of specialized hardware and cloud computing is extremely high. OpenAI claimed that the training cost for GPT-4 \citep{achiam2023gpt} exceeded 100 million dollars. The PaLM model developed by Google AI was trained for two months using 6144 TPUs for 10 million dollars \citep{chowdhery2023palm}. All this implies a need for faster and more cost-efficient optimization algorithms. It also suggests that early stopping \citep{prechelt2002early, bai2021understanding} in the training phase is a desirable feature whenever possible.

We focus in this work on computational efficiency during the training phase and consider the problem of unconstrained minimization of a loss function $\J \colon \mathbb{R}^p \to \mathbb{R}$, as follows
\begin{align}
    \label{eq:mainProblem}
    \min_{\theta \in \mathbb{R}^p} \J(\theta).
\end{align}
 
\paragraph{Continuous dynamical systems as optimization models.}
To achieve higher efficiency, it is necessary to deeply understand how algorithms work and how they relate to each other. A useful way to do this is by interpreting optimization algorithms as discrete versions of continuous dynamical systems \citep{ljung1977analysis}, further developed in \citep{harold1997stochastic, benaim2006dynamics, borkar2008stochastic, attouch2016fast, aujol2019optimal}. 
In deep learning, this approach is also quite fruitful; it has, in particular, been used to provide convergence proofs or further geometric insights \citep{davis2020stochastic, bolte2020conservative, barakat2021convergence, chen2023lion}.

In the spirit of \cite{castera2021inertial}, we consider the following continuous-time dynamical system introduced
in \cite{alvarez2002second} and referred to as DIN (standing for “dynamical inertial Newton”):
\begin{align}
\underbrace{\ddot\theta(t)}_{\text{\rm Inertial term}}+\underbrace{\alpha\,\dot\theta(t)}_{\text{\rm Friction term}}+\underbrace{\beta\,\nabla^{2}\J(\theta(t))\dot\theta(t)}_{\text{\rm Newtonian effects}}+\underbrace{\nabla \J(\theta(t))}_{\text{\rm Gravity effect}}= \ 0, \qquad t\geq 0,\label{eq:general_ode_din}
 \end{align} where $t$ is the time, $\J \colon \mathbb{R}^p \to \mathbb{R}$ is a loss function to be minimized (e.g., empirical loss in DL applications) as in \Cref{eq:mainProblem}, assumed $C^2$ with gradient $\nabla \J$ and Hessian $\nabla^2\J$. A key aspect of Equation \eqref{eq:general_ode_din} that places it between first- and second-order optimization is that a change of variables allows to describe it using only the gradient $\nabla \J$, since $\nabla^{2}\J(\theta(t))\dot\theta(t) = \frac{d}{dt} \nabla \J(\theta(t))$ (see Section \ref{sec:derivation} for details). This greatly reduces computational costs, as it can be discretized as a difference of gradients which does not require Hessian vector product, making it possible to design more practical algorithms, as shown in \cite{chen2019first, castera2021inertial, attouch2022first}.

We recover the continuous-time heavy ball system by assuming $\alpha>0$,  and removing the geometrical ``damping" term in Equation \eqref{eq:general_ode_din} through the choice $\beta=0$. A discrete version of this system corresponds to the Heavy Ball method  \citep{polyak1964some}, which is at the basis of SGD solvers with momentum in deep learning \citep{qian1999momentum, sutskever2013importance}. By allowing both $\alpha$ and $\beta$ to vary, we recover Nesterov acceleration \citep{nesterov1983method, su2016differential, attouch2019rate}.

\paragraph{Adaptive methods.}
Adaptive optimization methods, such as RMSprop \citep{tieleman2012lecture} and AdaGrad \citep{duchi2011adaptive}, modify the update dynamics by introducing coordinate-wise scaling of the gradient based on past information. These methods can be modeled by continuous-time ODEs of the following form, expressed here for the simple gradient system:
\begin{align}
    \label{eq:adaptiveMethods}
    \dot{\theta}(t) + \frac{1}{\sqrt{G(t,\theta(t)) + \epsilon}} \odot \nabla \mathcal{J}(\theta(t)) = 0, \quad t \geq 0,
\end{align}
where $\epsilon > 0$, $G(t,\theta(t)) \in \mathbb{R}^p$ represents accumulated  information. The scalar addition, square root, and division are understood coordinatewise and $\odot$ denotes the coordinate-wise product for vectors in $\mathbb{R}^p$. In AdaGrad or RMSprop, $G(t,\theta(t))$ is a gradient amplitude averaged of the  form:
\begin{align}
    \label{eq:accumulationGradient}
    G(t,\theta(t)) := \int_0^t \nabla \mathcal{J}(\theta(\tau))^2 \, d \mu_t(\tau),
\end{align}
for different choices of $\mu_t$ --- uniform for AdaGrad and moving average for RMSprop. This generally improves performance, see the pioneering work \citep{duchi2011adaptive, tieleman2012lecture}.

\paragraph{Our approach.}
We combine the ``dynamical inertial Newton" method (DIN)  from Equation \eqref{eq:general_ode_din} with an RMSprop adaptive gradient scaling. This allows us to take into account second-order information for the RMSProp scaling. Computationally, this second-order information is expressed using a time derivative. In discrete time, this will provide a second-order intelligence with the same computational cost as gradient evaluation. The resulting continuous time ODE is given as follows:
\begin{align}
    \label{eq:dynRMSprop}
    &\ddot\theta(t) + \alpha\,\dot\theta(t) + \beta\,\frac{d}{dt} \rmsprop(\mathcal{J}(\theta(t))) + \rmsprop(\mathcal{J}(\theta(t))) = 0,  \qquad\,t \geq 0 \\ 
    &\text{where }\rmsprop(\mathcal{J}(\theta(t))) = \frac{1}{\sqrt{G(t,\theta(t)) + \epsilon}} \odot \nabla \mathcal{J}(\theta(t)) \nonumber
\end{align}
with $G$ of the form \eqref{eq:accumulationGradient} with an adequate time-weight distribution $\mu_t$ corresponding to the RMSProp scaling. A discretization of this continuous time system, combined with careful memory management, results in our new optimizer INNAprop, see  Section \ref{sec:thealgorithm}. %\jer{donner lien vers l'algo avant sa "derivation"} Section \ref{sec:derivation} %for details of our derivation.

\paragraph{Relation with existing work.}

To improve the efficiency of stochastic gradient descent (SGD), two primary strategies are used: leverage local geometry for having clever directions and incorporate momentum to accelerate convergence. These approaches include accelerated methods (e.g., Nesterov’s acceleration \citep{nesterov1983method, dozat2016incorporating}, momentum SGD \citep{polyak1964some,qian1999momentum,sutskever2013importance}, and adaptive methods (e.g., Adagrad \citep{duchi2011adaptive}, RMSProp \citep{tieleman2012lecture}), which adjust learning rates per parameter.

Adam remains the dominant optimizer in deep learning. It comes under numerous variants proposed to improve its performance or to adapt it to specific cases  \citep{dozat2016incorporating, shazeer2018adafactor, reddi2019convergence, loshchilov2017decoupled, zhuang2020adabelief}. Adafactor \citep{shazeer2018adafactor} improves memory efficiency, Lamb \citep{you2019large} adds layerwise normalization, and Lion \citep{chen2023symbolic} uses sign-based momentum updates. AdEMAMix \citep{pagliardini2024ademamix} combines two EMAs, while Defazio et al. \citep{defazio2024road} introduced a schedule-free method incorporating Polyak-Ruppert averaging with momentum.

One of the motivations of our work is the introduction of second-order properties in the dynamics akin to Newton's method. Second-order optimizers are computationally expensive due to frequent Hessian computations \citep{gupta2018shampoo, martens2015optimizing}. Their adaptation to large scale learning settings require specific developments \citep{jahani2021doubly, qian2021basis}. For example, the Sophia optimizer \citep{liu2023sophia}, designed for large language models, uses a Hessian-based pre-conditioner to penalize high-curvature directions. In this work, we draw inspiration from  INNA  \citep{castera2021inertial}, based on the continuous time dynamics introduced by \citep{alvarez2002second}, which combines gradient descent with a Newtonian mechanism for first-order stochastic approximations. %They do not however use local geometry adaption.

Our proposed method, INNAProp, integrates the algorithm INNA, which features a Newtonian effect with cheap computational cost, with the gradient scaling mechanism of RMSprop. This  preserves the efficiency of second-order methods and the adaptive features of RMSprop while significantly reducing the computational overhead caused by Hessian evaluation. Specific hyperparameter choices for our method allow us to recover several existing optimizers as special cases.

\paragraph{Contributions.}
They can be summarized as follows:
\begin{itemize}[label=\small$\bullet$, left=0.2cm]
    \item We introduce INNAprop, a new optimization algorithm that combines the Dynamical Inertial Newton (DIN) method with RMSprop’s adaptive gradient scaling, efficiently using second-order information for large-scale machine learning tasks.
    We obtain a second-order optimizer with computational requirements similar to first-order methods like AdamW, making it suitable for deep learning (see Section \ref{sec:derivation} and Appendix \ref{sec:detailsinnaprop}).
    \item We provide a continuous-time explanation of INNAprop, connecting it to second-order ordinary differential equations (see \Cref{sec:ourmethod} and \Cref{eq:dynRMSprop}). We discuss many natural possible discretizations and show that INNAprop is empirically the most efficient. Let us highlight a key feature of our method: it incorporates second-order terms in space without relying on Hessian computations or inversions of linear systems which are both prohibitive in deep learning.
    
    \item We show through extensive experiments that INNAprop matches or outperforms AdamW in both training speed and final accuracy on benchmarks such as image classification (CIFAR-10, ImageNet) and language modeling (GPT-2) (see Section \ref{sec:experiments}).
\end{itemize}

We describe our algorithm and its derivation in Section \ref{sec:ourmethod}. Hyperparameter tuning recommendations and our experimental results are provided in Section \ref{sec:experiments}.

\section{INNAprop: a second-order method in space and time based on RMSProp}
\label{sec:ourmethod}
\subsection{The algorithm}
\label{sec:thealgorithm}

Our method is built on the following \Cref{alg:INNAprop}, itself derived from a combination of INNA 
\citep{castera2021inertial} and RMSprop \citep{tieleman2012lecture} (refer to Section \ref{sec:derivation} for more details). The following version of the method is the one we used in all experiments. It includes the usual ingredients of deep-learning training: mini-batching, decoupled weight-decay, and scheduler procedure. For a simpler, ``non-deep learning" version, refer to Algorithm \ref{alg:innaprop_brut} in Appendix \ref{sec:detailsinnaprop}.

\begin{algorithm}
\caption{Deep learning implementation of $\INNAprop$ }
\begin{algorithmic}[1]
\STATE \textbf{Objective function:} $\J(\theta) = \frac{1}{n} \sum_{n = 1}^N \J_{n}(\theta)$ for $\theta \in \R^p$.
\STATE \textbf{Learning step-sizes:} $ \gamma_k := \{\text{SetLrSchedule}(k)\}_{k \in \NN}$ where $\gamma_0$ is the initial learning rate.
\STATE{\textbf{Hyper-parameters:} $\sigma \in [0,1]$, $\alpha \geq 0$, $\beta > \sup_{k \in \mathbb{N}} \gamma_k$, $\lambda \geq 0$, $\epsilon = 10^{-8}$.} 
\STATE \textbf{Mini-batches:} $(\mb_k)_{k \in \N}$ of nonempty subsets of $\{1,\ldots,N\}$.

\STATE \textbf{Initialization:}  time step $k \leftarrow 0$, parameter vector $\theta_0$, $v_0 = 0$, $\psi_0 = (1 - \alpha \beta) \theta_0$.
\FOR{$k=1$ {\bfseries to} K}
    \STATE $\bm{g}_k = \frac{1}{\lvert \mb_k \rvert}\sum_{n\in \mb_k} \nabla\J_n(\bm{\theta}_k)$ \COMMENT{select batch $\mb_k$ and return the corresponding gradient}
    \STATE{$\gamma_k \leftarrow \text{SetLrSchedule}(k)$} \COMMENT{see above and \Cref{remark:wp}}
    \STATE $\bm{\theta}_k \gets (1 - \lambda \gamma_k) \bm{\theta}_k $ \COMMENT{decoupled weight decay}
    \STATE $\bm{v}_{k+1} \gets \sigma \bm{v}_{k} + (1-\sigma) \bm{g}_k^2$ 
    \STATE $\bm{\hat{v}}_{k+1} \gets \bm{v}_{k+1}/(1 - \sigma^k)$
    \STATE $\bm{\psi}_{k+1} \gets \left(1 - \frac{\gamma_k}{\beta}\right)\bm{\psi}_k + \gamma_k \left(\frac{1}{\beta} - \alpha\right) \bm{\theta}_k$
    \STATE $\bm{\theta}_{k+1} \gets \left( 1 +  \frac{\gamma_k  (1 - \alpha\beta)}{\beta - \gamma_k} \right) \bm{\theta}_k  - \frac{\gamma_k}{\beta - \gamma_k} \bm{\psi}_{k+1} - \gamma_k\beta \left (\bm{g}_k / (\sqrt{\bm{\hat{v}}_{k+1}} + \epsilon \right)$

\ENDFOR
\RETURN{$\bm{\theta}_{K+1}$}
\end{algorithmic}
\label{alg:INNAprop}
\end{algorithm}

In \Cref{alg:INNAprop}, \text{SetLrSchedule} is the ``scheduler" for step-sizes; it is defined as a custom procedure for handling learning rate sequences for different networks and databases. To provide a full description of our algorithm, we provide detailed explanations of the scheduler procedures used in our experiments (Section \ref{sec:experiments}) in Appendix \ref{sec:schedulers}, along with the corresponding benchmarks.

\begin{remark}[Well posedness]{\rm 
Observe that, for all schedulers $\gamma_k < \beta$ for $k \in \mathbb{N}$, so that $\INNAprop$ is well-posed (line 13 in Algorithm \ref{alg:INNAprop}, the division is well defined).}
\label{remark:wp}
\end{remark}

\subsection{Derivation of the algorithm}
\label{sec:derivation}
There are several ways to combine $\rmsprop$ and INNA, or DIN its second-order form, as there exist several ways to do so with the heavy ball method and $\rmsprop$. We opted for the approach below because of its mechanical and geometrical appeal and its numerical success (see Appendix \ref{sec:detailsinnaprop} for further details). 
Consider the following dynamical inertial Newton method \citep{alvarez2002second}:
 \begin{align}
 \label{eq:physicalIntuitionSmooth}
\ddot\theta(t)+\alpha\,\dot\theta(t)+\beta\,\frac{d}{dt}\nabla\J(\theta(t))+ \nabla \J(\theta(t))= \ 0, \quad t \geq 0,
 \end{align}
%\ed{tu peux faire reference à l'intro ici}
as in \Cref{eq:general_ode_din} and replacing  $\nabla^{2}\J(\theta(t))\dot\theta(t)$ by $\frac{d}{dt} \nabla \J(\theta(t))$. Using finite differences with a fixed time step $\gamma$ for discretization, replacing in particular the gradient derivatives by gradient differences:
$$\frac{d}{dt}\nabla\J(\theta(t)) \simeq \frac{\nabla\J(\theta_{k+1})-\nabla\J(\theta_k)}{\gamma}, $$%\hspace{0.5cm} \dot{\theta(t)} \simeq \frac{\theta_{k+1} - \theta_k}{\gamma} $$
where  
$\theta_k,\theta_{k+1}$ correspond to two successive states around the time $t$.

Setting $\nabla\J(\theta_{k}) = g_k$, we obtain $\displaystyle \frac{\theta_{k+1} - 2\theta_k + \theta_{k-1}}{\gamma} + \alpha \frac{\theta_{k} - \theta_{k-1}}{\gamma} + \beta \frac{{g_k} - g_{k-1}}{\gamma} + g_{k-1} = 0.   
$

%\paragraph{Embedding $\rmsprop$ directly in DIN:}  
%\begin{align}
 %\label{eq:dinrmsprop_continous}
 %\underbrace{\ddot\theta(t)}_{\text{\rm Inertial term}}+\underbrace{\alpha\,\dot\theta(t)}_{\text{\rm Friction term}}+\underbrace{\beta\,\frac{d}{dt}\rmsprop (\nabla \J(\theta(t)))}_{\text{\rm Newtonian effects}}+\underbrace{ \rmsprop(\nabla \J(\theta(t)))}_{\text{\rm Gravity effect}}= \ 0, \quad \text{for $t\in [0,+\infty)$},
 %\end{align}
To provide our algorithm with an extra second-order geometrical intelligence, we use the proxy of $\rmsprop$ direction in place of the gradient. 

Choose $\sigma > 0$ and $\epsilon > 0$, and consider:
\begin{align}
& v_{k+1} = \sigma v_{k} + (1 - \sigma) g_{k}^2 \label{v}\\
&\frac{\theta_{k+1} - 2\theta_k + \theta_{k-1}}{\gamma} + \alpha \frac{\theta_{k} - \theta_{k-1}}{\gamma} + \beta \frac{\frac{g_k}{\sqrt{v_{k+1}} + \epsilon} - \frac{g_{k-1}}{\sqrt{v_k} + \epsilon}}{\gamma} + \frac{g_{k-1}}{\sqrt{v_k} + \epsilon} = 0.
\label{eq:dinrmsprop_start}
\end{align}
Although this system has a natural mechanical interpretation, its memory footprint is abnormally important for this type of algorithm: for one iteration of the system \eqref{v}-\eqref{eq:dinrmsprop_start}, it culminates at 6 full dimension memory slots, namely $g_{k-1}$, $g_k$, $\theta_{k-1}$, $\theta_{k}$, $v_k$, and $v_{k+1}$ before the evaluation of \eqref{eq:dinrmsprop_start}.

Therefore, we proceed to rewrite the algorithm in another system of coordinates. The computations and the variable changes are provided in Appendix \ref{sec:detailsinnaprop}. We eventually obtain: 
\begin{align*}
& v_{k+1} = \sigma v_{k} + (1 - \sigma) g_{k}^2 \\
\psi_{k+1} &= \psi_k \left(1 - \frac{\gamma}{\beta}\right) + \gamma \left(\frac{1}{\beta} - \alpha\right) \theta_k, \\
\theta_{k+1} & = \left(  1 + \frac{\gamma  (1 - \beta\alpha)}{\beta - \gamma } \right) \theta_k   - \frac{\gamma}{\beta - \gamma } \psi_{k+1} - \gamma\beta \frac{g_k}{\sqrt{v_{k+1}} + \epsilon}
\end{align*}
which only freezes 3 full dimension memory slots corresponding to $v_{k}$, $\psi_{k}$, $\theta_k$. As a result, the memory footprint is equivalent to that of the Adam optimizer (see \Cref{table:updaterules}).

\begin{remark}[On other possible discretizations]\label{remark:discret}{\rm 
(a) If we use the proxy of RMSprop directly with INNA \citep{castera2021inertial}, we recover indeed INNAprop through a rather direct derivation (see Appendix \ref{sec:alternative_innaprop} for more details). Our motivation to start from the ``mechanical" version of the algorithm is to enhance our understanding of the geometrical features of the algorithm.\\
(b) RMSprop with momentum \citep{graves2013generating} is obtained by a discretization of the heavy ball continuous time system, using a momentum term and an RMSprop proxy. It would be natural to proceed that way in our case, and it indeed leads to a different method (see Appendix \ref{sec:innaprop_momentum}). However, the resulting algorithm appears to be numerically unstable (see Figure \ref{fig:innaprop_momentum} for an illustration). \\ %\ed{what does it mean ill posed?}.
(c) Incorporating $\rmsprop$ as it is done in $\adam$ using momentum leads to a third method (see Appendix \ref{sec:dinadam}), which appears to be extremely similar to NAdam \citep{dozat2016incorporating}; it was thus discarded.}
\end{remark}

\begin{remark}[A family of algorithms indexed by $\alpha,\beta$]\label{remark:family}{\rm $\INNAprop$ can be seen as a family of methods indexed by the hyperparameters $\alpha$ and $\beta$. When $\beta=0$, we recover a modified version of $\rmsprop$ with momentum \citep{graves2013generating} (see Appendix \ref{sec:particularcase}). For $\alpha=\beta=1$, $\INNAprop$ with its default initialization, boils down to AdamW without momentum ($\beta_1 = 0$), see Appendix \ref{sec:particularcase} and \Cref{table:updaterules}. By setting $\alpha=\beta=1$, we empirically recover the behavior of AdamW. Experiments demonstrate that this consistently aligns with AdamW, suggesting that AdamW can be seen as a special case within the broader INNAprop family. See Appendix \ref{sec:particularcase} for further details and illustrations.} We now explain how these hyperparameters ($\alpha,\beta$) have been tuned on ``small size" problems.
\end{remark}

\section{Empirical evaluation of INNAprop}
\label{sec:experiments}
We conduct extensive comparisons of the proposed algorithm and the AdamW optimizer, which is dominantly used in image classification \citep{chen2018closing, zhuang2020adabelief, touvron2021training, mishchenko2023prodigy} and language modeling tasks \citep{brown2020language, hu2021lora, liu2023sophia}. Hyperparameter tuning \citep{sivaprasad2020optimizer} is a crucial issue for this comparison, and we start with this. As a general rule, we strive to choose the hyperparameters that give a strong baseline for AdamW (based on literature or using grid search). Unless stated differently, our experiments use the AdamW optimizer \footnote{\url{https://pytorch.org/docs/stable/generated/torch.optim.AdamW.html}} with its default settings as defined in widely-used libraries \citep{NEURIPS2019_9015, jax2018github, 45381}: $\beta_1=0.9$, $\beta_2=0.999$, $\lambda = 0.01$ and $\epsilon=1e-8$. For INNAprop, unless otherwise specified, the default settings for the RMSprop component align with those of AdamW: $\sigma=0.999$ and $\epsilon=1e-8$. 

The INNAprop method and the AdamW optimizer involve different classes of hyperparameters; some of them are common to both algorithms, and some are specific. Our hyperparameter tuning strategy for both algorithms is summarized in Table \ref{tab:hyperparmTuningStartegy}. 

We begin this section with the tuning of parameters $\alpha,\beta$ for INNAprop on CIFAR10 with VGG and ResNet architectures and then use these parameters on larger datasets and models. We use as much as possible the step size scheduler and weight decay settings reported in the literature for the AdamW optimizer, which we believe to be well-adjusted and provide adequate references for each experiment. These are used both for AdamW and INNAprop. With this protocol, we only perform minimal hyperparameter tuning for INNAprop for larger-scale experiments. This is due to constrained computational resources. We aim to demonstrate the typical performance of the Algorithm \ref{alg:INNAprop}, rather than its peak performance with extensive tuning. 

\begin{table}[!ht]
    \begin{center}
        \caption{\footnotesize Hyperparameter tuning strategy for INNAprop and AdamW: AdamW is systematically favored.}
        \label{tab:hyperparmTuningStartegy}
        \setlength{\tabcolsep}{4pt}
        \renewcommand{\arraystretch}{1.1}
        {\fontsize{9}{11}\selectfont{
        \begin{tabular}{l|c|c|c}
        \toprule
        \textbf{Parameters} & \textbf{AdamW tuning} & \textbf{INNAprop tuning} & \textbf{Comparative advantage} \\ \midrule
        Learning rate & Literature or grid search tuning & Reused from AdamW & AdamW favored \\ \hline
        Step size scheduler & Literature & Reused from AdamW & N/A \\ \hline
        Weight decay & Literature or grid search tuning & Reused from AdamW & AdamW favored \\ \hline
        RMSprop parameter & Default or literature & Reused from AdamW & AdamW favored \\ \hline
        Inertial parameters ($\alpha, \beta$) & N/A & Tuned on CIFAR-10 & N/A \\ 
        \bottomrule
        \end{tabular}
        }}
    \end{center}
\end{table}

\subsection{Tuning INNAprop on CIFAR-10 with VGG11 and ResNet18}
\label{sec:tuneinnaprop}

\begin{figure}[h]
    \centering
    \begin{subfigure}[b]{0.8\textwidth}
        \begin{minipage}{0.2\textwidth} % Ajuster la largeur si nécessaire
            \caption{20 epochs}
        \end{minipage}%
        \begin{minipage}{0.7\textwidth}
            \includegraphics[width=\textwidth]{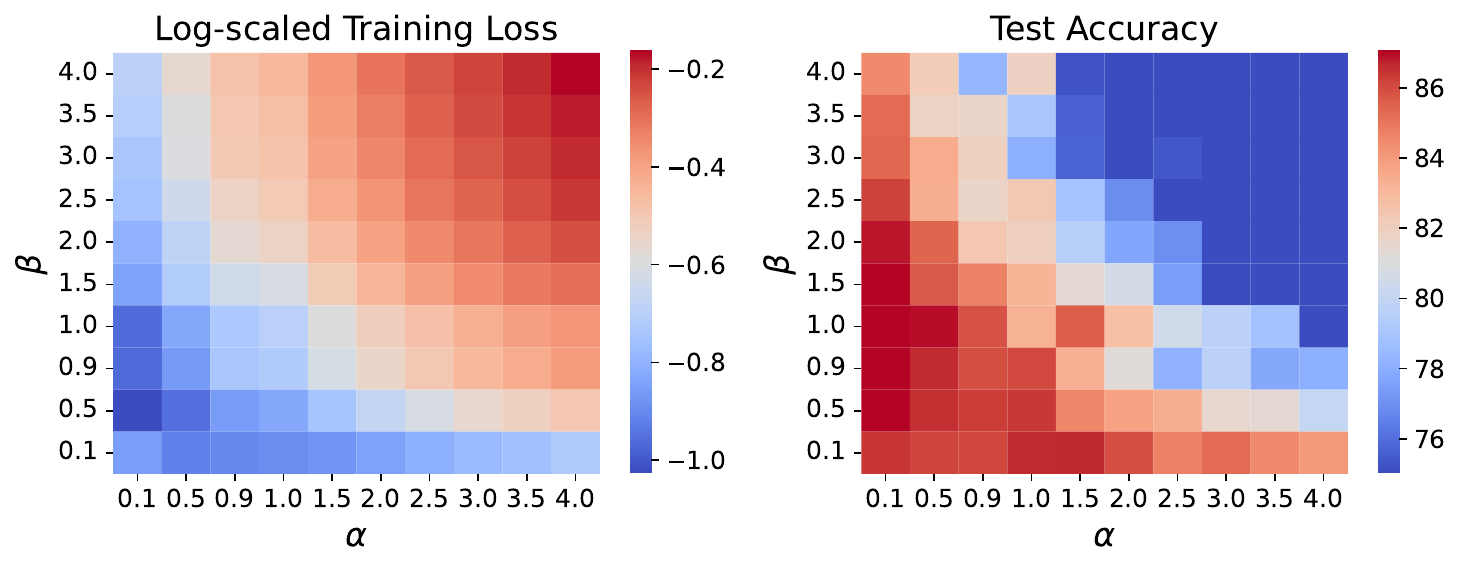}
        \end{minipage}
    \end{subfigure}
    
    \vspace{0.3cm} % Espacement vertical entre les sous-figures

    \begin{subfigure}[b]{0.8\textwidth}
        \begin{minipage}{0.2\textwidth} % Ajuster la largeur si nécessaire
            \caption{200 epochs}
        \end{minipage}%
        \begin{minipage}{0.7\textwidth}
            \includegraphics[width=\textwidth]{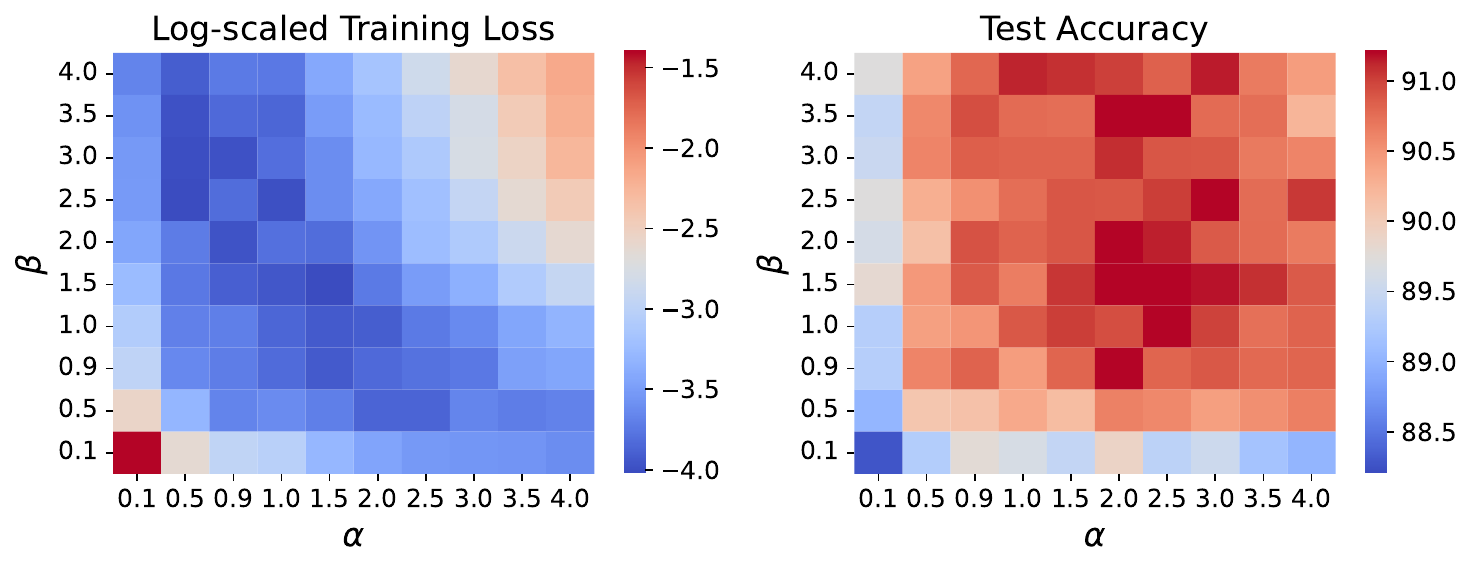}
        \end{minipage}
    \end{subfigure}
    
    \caption{\footnotesize Log-scale training loss and test accuracies for hyperparameters $(\alpha, \beta)$ with VGG11 on CIFAR10 at 20 and 200 epochs. Optimal learning rate $\gamma_0 = 10^{-3}$ and weight decay $\lambda = 0.01$, with one random seed.}
    \label{fig:heatmaps_vgg_wd_0.01}
\end{figure}

\paragraph{Hyperparameter tuning:} We tune ($\alpha, \beta$) using VGG11 \citep{simonyan15very} and ResNet18 \citep{he2016deep} models trained on CIFAR10 \citep{krizhevsky2010cifar}, together with the initial learning rate $\gamma_0$ to ensure proper training. We fix a cosine scheduler where $T_{\text{max}} = 200$ and $\gamma_{\min} = 0$ (see \Cref{sec:schedulers} for more details) and we consider two weight decay parameters $\lambda = 0$ or $\lambda = 0.01$. Our experiment suggests using an initial learning rate $\gamma_0 = 10^{-3}$, which is the baseline value reported for AdamW in this experiment (see \Cref{sec:hyperparameterTuning}). For $\INNAprop$, we optimize only the hyperparameters $\alpha$ and $\beta$, using test accuracy and training loss as the optimization criteria. A grid search is performed over $(\alpha, \beta) \in \{0.1, 0.5, 0.9,\dots, 3.5, 4.0\}$ using \texttt{optuna} \citep{akiba2019optuna}.  In Figure \ref{fig:heatmaps_vgg_wd_0.01}, we detail the obtained training loss and test accuracy for various $(\alpha, \beta)$ configurations over short training durations (20 epochs) and long training durations (200 epochs) for VGG11 with weight decay $\lambda=0.01$. Our criteria (short and long training duration) are chosen to find parameters $(\alpha,\beta)$ that provide a rapid decrease in training loss in the early stages and the best test accuracy for long training duration.

These results highlight a tendency for efficient couples; we choose for further experiments the values $(\alpha,\beta)=(0.1,0.9)$ which correspond to aggressive optimization of the training loss for short training durations, and $(\alpha,\beta)=(2.0,2.0)$ which provides very good results for longer training durations. Additional results for VGG11 and ResNet18 with and without weight decay are in Appendix \ref{sec:heatmaps}, which are qualitatively similar.

\begin{figure}[!ht]
    \centering
    \includegraphics[width=\textwidth]{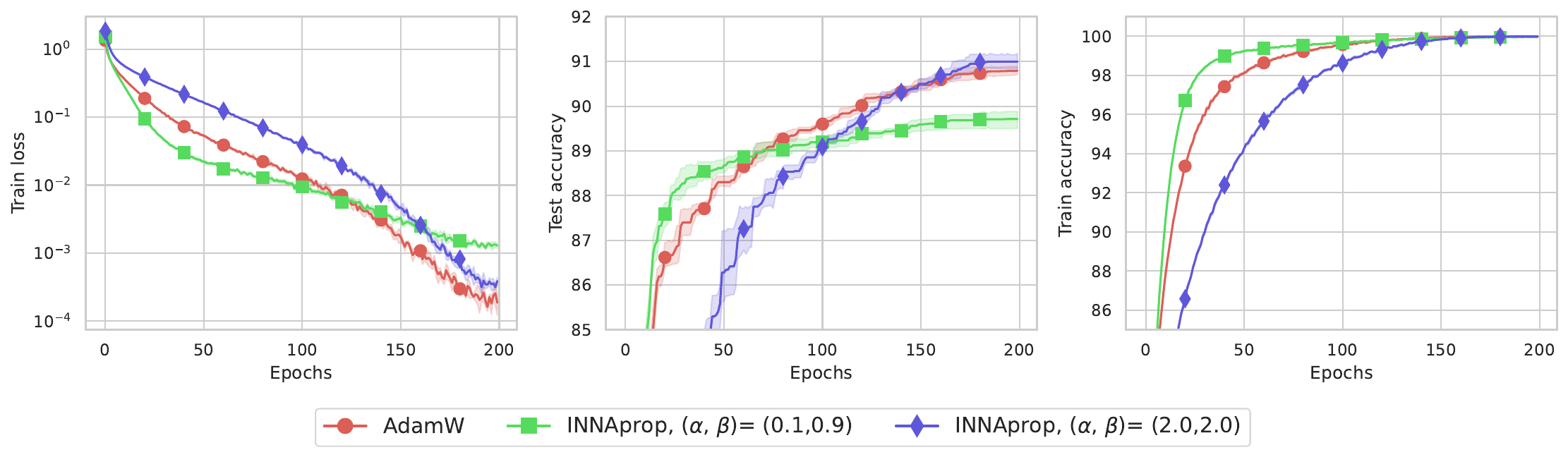}
    \caption{\footnotesize Training VGG11 on CIFAR10. Left: train loss, middle: test accuracy (\%), right: train accuracy (\%), with 8 random seeds.}
    \label{fig:vgg11_cifar10}
\end{figure}

\paragraph{Validation and comparison with AdamW:} We confirm our hyperparameter choices ($\gamma_0 = 10^{-3}$, $\lambda = 0.01$) by reproducing the experiment with 8 random seeds and comparing with AdamW using the same settings. Based on hyperparameter tuning, we select two pairs of $(\alpha, \beta)$ with different training speeds. As shown in Figure \ref{fig:vgg11_cifar10} (and \Cref{sec:additionalExperiments} for ResNet18), with $(\alpha, \beta) = (0.1, 0.9)$, INNAprop improves training loss and test accuracy rapidly by the 100th epoch, maintaining the highest training accuracy. With $(\alpha, \beta) = (2.0, 2.0)$, INNAprop trains more slowly but achieves higher final test accuracy. This is aligned with the experiments described in \Cref{fig:heatmaps_vgg_wd_0.01}. In \Cref{tab:cifar10}, we compare the performance of different networks on CIFAR-10 using INNAprop and AdamW optimizers.

\begin{table*}[h!]
    \begin{center}
        \caption{\footnotesize Test accuracy (\%) of ResNet-18, VGG11, and DenseNet121 on CIFAR-10 using AdamW optimized weight decay and learning rate. Results are averaged over eight runs.}
        \label{tab:cifar10}
        \setlength{\tabcolsep}{4pt}
        \renewcommand{\arraystretch}{1.1}
        {\fontsize{8}{10}\selectfont{
    \begin{tabular}{c|c|c}
        \toprule
        \textbf{Model} & \textbf{Optimizer} & \textbf{Test accuracy} \\ 
        \midrule
        \multicolumn{3}{c}{Training on CIFAR-10 over 200 epochs} \\
        \midrule
        \multirow{2}{*}{ResNet18} & AdamW & 91.14 \\
         & INNAprop ($\alpha=2.0, \beta=2.0$) & \textbf{91.58} \\
        \midrule
        \multirow{2}{*}{VGG11} & AdamW & 90.79 \\
         & INNAprop ($\alpha=2.0, \beta=2.0$) & \textbf{90.99} \\
        \midrule
        \multirow{2}{*}{DenseNet121} & AdamW & 86.19 \\
         & INNAprop ($\alpha=0.1, \beta=0.9$) & \textbf{86.91} \\
        \bottomrule
    \end{tabular}
        }}
    \end{center}
\end{table*}

\begin{remark}[Trade-off between fast learning and good generalization]
{\rm For CIFAR-10 experiments, INNAprop offers flexibility in adjusting convergence speed through $(\alpha, \beta)$. Faster training configurations generally lead to weaker generalization compared to slower ones, highlighting the trade-off between quick convergence and generalization \citep{wilson2017marginal, zhang2020adaptive}.}
\label{remark:tradeoff}
\end{remark}

\subsection{Extensive experiments on large-scale vision models}
\label{sec:imagexp}
We present experiments on large-scale vision benchmarks with the hyperparameters of Section \ref{sec:tuneinnaprop}.

\paragraph{Resnets on ImageNet:} We consider the larger scale ImageNet-1k benchmark \citep{krizhevsky2012imagenet}. We train a ResNet-18 and a ResNet-50 \citep{he2016deep} for 90 epochs with a mini-batch of size of 256 as in \cite{chen2023symbolic, zhuang2020adabelief}. We used the same cosine scheduler for both AdamW and INNAprop with initial learning rate $\gamma_0 = 10^{-3}$ as reported in \cite{chen2023symbolic, zhuang2020adabelief, chen2018closing}. The weight decay of AdamW is set to $\lambda=0.01$ for the ResNet18,  instead of $\lambda=0.05$ reported in \cite{zhuang2020adabelief, chen2018closing} because it improved the test accuracy from $67.93$ to $69.43$. The results of the ResNet18 experiment are presented in \Cref{fig:ResNetAppendix} in \Cref{sec:additionalExperiments}. The figure shows that our algorithm with $(\alpha, \beta) = (0.1, 0.9)$ outperforms AdamW in test accuracy (70.12 vs 69.34), though the training loss decreases faster initially but slows down towards the end of training.

\begin{figure}[!ht]
    \centering
    \includegraphics[width=\textwidth]{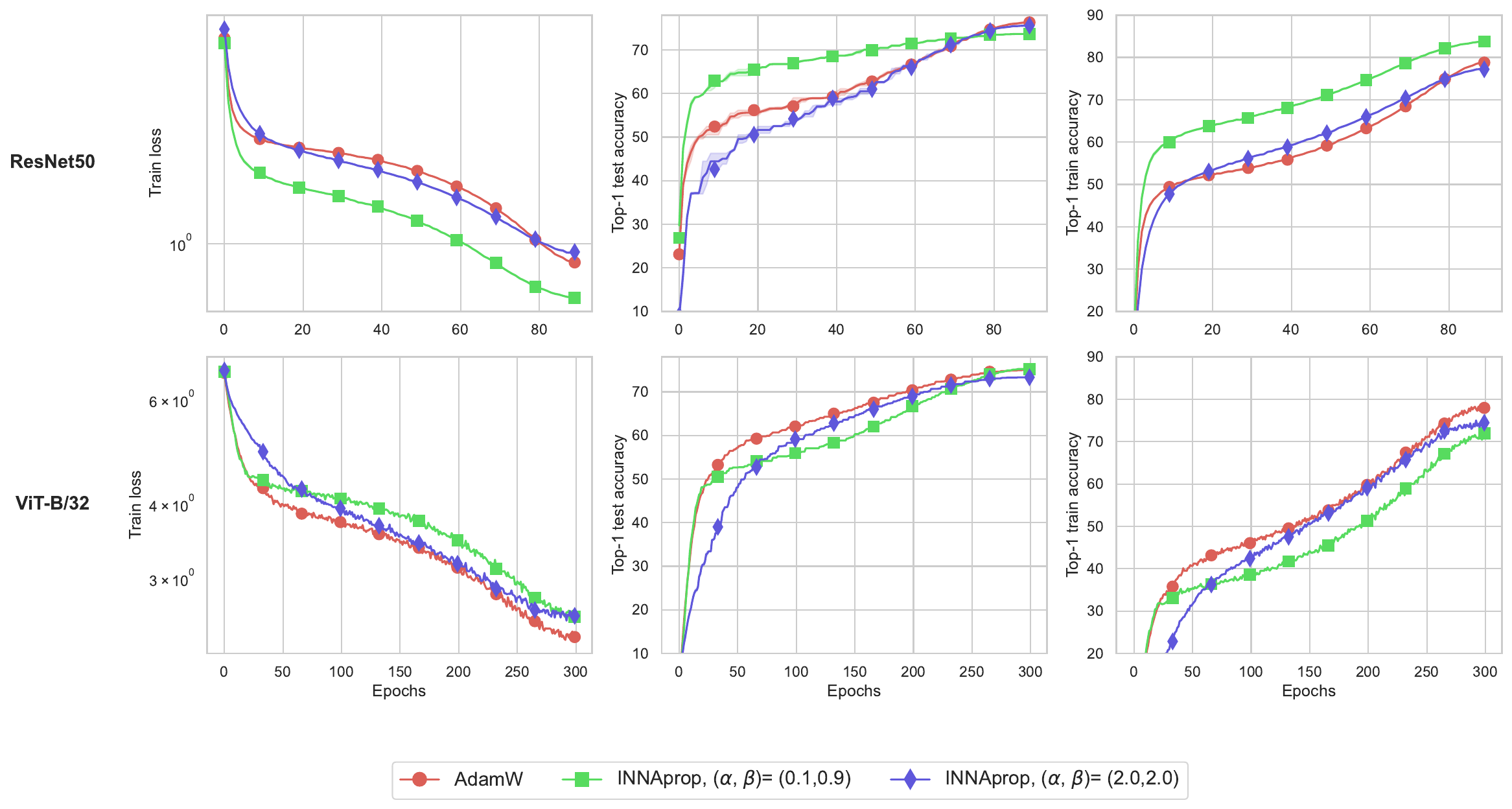}
    \caption{\footnotesize Training a ResNet50 (top) and ViT-B/32 (bottom) on ImageNet. Left: train loss, middle: Top-1 test accuracy (\%), right: Top-1 train accuracy (\%). 3 random seeds.}
    \label{fig:ResNetVit}
\end{figure}

For the ResNet50, we kept the value $\lambda = 0.1$ as reported in \cite{zhuang2020adabelief, chen2018closing}. For INNAprop, we tried two weight decay values $\{0.1, 0.01\}$ and selected $\lambda = 0.01$ as it resulted in a faster decrease in training loss. We report the results in \Cref{fig:ResNetVit}, illustrating the advantage of INNAprop. As noted in Section \ref{sec:tuneinnaprop}, INNAprop with $(\alpha, \beta) = (0.1, 0.9)$ reduces training loss quickly but has lower test accuracy compared to AdamW or INNAprop with $(\alpha, \beta) = (2.0, 2.0)$. For $(\alpha, \beta) = (2.0, 2.0)$, the loss decrease is similar to AdamW, with no clear advantage for either method. This obviously suggests developing scheduling strategies for damping parameters $(\alpha, \beta)$. This would require a much more computation-intensive tuning, far beyond the numerical resources used in the current work. In \Cref{tab:imagenet}, we present the performance of INNAprop achieved using minimal hyperparameter tuning, as explained in \Cref{tab:hyperparmTuningStartegy}.

\paragraph{Vision transformer (ViT) on ImageNet:} We performed the same experiment with a ViT-B/32 architecture over 300 epochs with a mini-batch size of 1024, following \cite{defazio2023learning, mishchenko2023prodigy}. For AdamW, we used a cosine scheduler with a linear warmup (30 epochs) and the initial learning rate and weight decay from \cite{defazio2023learning}. For INNAprop, we tested weight decay values of $\{0.1, 0.01\}$, selecting $\lambda = 0.1$ for better test accuracy. Results in \Cref{fig:ResNetVit} show the advantage of INNAprop. For faster convergence using INNAprop $(0.1,0.9)$, we recommend a weight decay of $\lambda=0.01$ (see Figure \ref{fig:ViTAppendix} in the Appendix).

In the ImageNet experiments, we evaluated INNAprop for rapid early training and optimal final test accuracy without tuning $(\gamma_0, \alpha, \beta)$. For ViT-B/32 with $\lambda=0.1$, INNAprop achieved lower training loss and higher final test accuracy than AdamW (75.23 vs. 75.02). %For ResNet-18, INNAprop also outperformed AdamW in final test accuracy with $\lambda=0.01$ (70.12 vs. 69.34).

\begin{table*}[h!]
    \begin{center}
        \caption{ \footnotesize Top-1 and Top-5 accuracy (\%) of ResNet-18, ResNet-50, and ViT-B/32 on ImageNet. Results are averaged from three runs for ResNets and one run for ViT-B/32. AdamW favored as in \Cref{tab:hyperparmTuningStartegy}.} 
        \label{tab:imagenet}
        \setlength{\tabcolsep}{4pt}
        \renewcommand{\arraystretch}{1.1}
        {\fontsize{8}{10}\selectfont{
    \begin{tabular}{c|c|c|c}
        \toprule
        \textbf{Model} & \textbf{Optimizer} & \textbf{Top-1 accuracy} & \textbf{Top-5 accuracy} \\ 
        \midrule
        \multicolumn{4}{c}{Train from scratch on ImageNet} \\
        \midrule
        \multirow{2}{*}{ResNet18} & AdamW & 69.34 & 88.71 \\
         & INNAprop ($\alpha=0.1, \beta=0.9$) & \textbf{70.12} & \textbf{89.21} \\
        \midrule
        \multirow{2}{*}{ResNet50} & AdamW & 76.33 & 93.04 \\
         & INNAprop ($\alpha=1.0, \beta=1.0$) & \textbf{76.43} & \textbf{93.15} \\
        \midrule
        \multirow{2}{*}{ViT-B/32} & AdamW & 75.02 & 91.52 \\
         & INNAprop ($\alpha=0.1, \beta=0.9$) & \textbf{75.23} & \textbf{91.77} \\
        \bottomrule
    \end{tabular}
        }}
    \end{center}
\end{table*}

\begin{figure}[!ht]
    \centering
    \includegraphics[width=\textwidth]{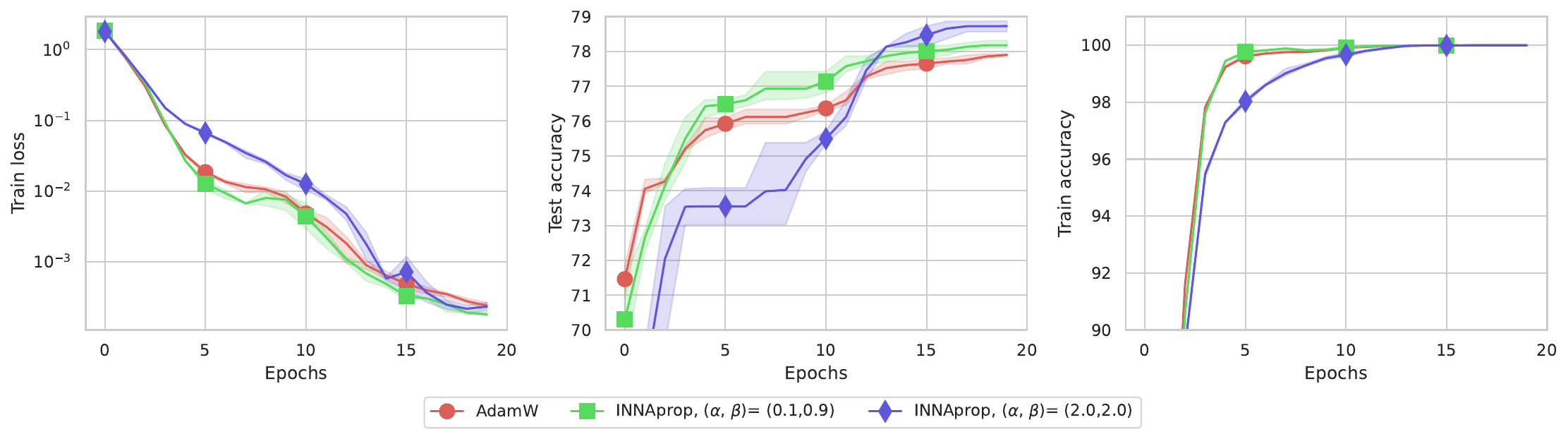}
    \caption{\footnotesize Finetuning a VGG11 on Food101. Left: train loss, middle: test accuracy (\%), right: train accuracy (\%). Qualitatively similar results for ResNet18 are in \Cref{fig:food101appendix} in \Cref{sec:additionalExperiments}. 3 random seeds.}
    \label{fig:food101}
\end{figure}

\paragraph{Fintetuning VGG11 and ResNet18 models on Food101:}
We fine-tuned ResNet-18 and VGG-11 models on the Food101 dataset \citep{bossard2014food} for 20 epochs, using pre-trained models on ImageNet-1k. Since weight decay and learning rate values for AdamW were not found in the literature, we chose the default AdamW weight decay value, $\lambda = 0.01$. We used a cosine scheduler and tried one run for each initial learning rate value in $\{10^{-5}, 5 \times 10^{-5}, 10^{-4}, 5 \times 10^{-4}, 10^{-3}\}$. The best result for AdamW was obtained for $\gamma_0 = 10^{-4}$, and we kept the same setting for INNAprop. See for this \Cref{fig:food101}, where INNAprop performs no worse than AdamW on three random seeds. 

\paragraph{Conclusion and recommendation for image classification:}
Tuning $(\alpha, \beta)$ significantly impacts training. Based on heatmaps in Section \ref{sec:tuneinnaprop} and figures in Section \ref{sec:imagexp}, we recommend using $\alpha = 0.1$ and $\beta \in [0.5, 1.5]$ for shorter training (e.g., fine-tuning). For longer training, $\alpha, \beta \geq 1$ is preferable. In both cases, our algorithm either matches or outperforms AdamW.

\subsection{Pre-training and fine-tuning GPT2}
\label{sec:llmxp}
We present experimental results on LLMs using the hyperparameters selected as in Section \ref{sec:tuneinnaprop}.

\begin{figure}[ht!]
\centering
\centering
\includegraphics[width=\textwidth]{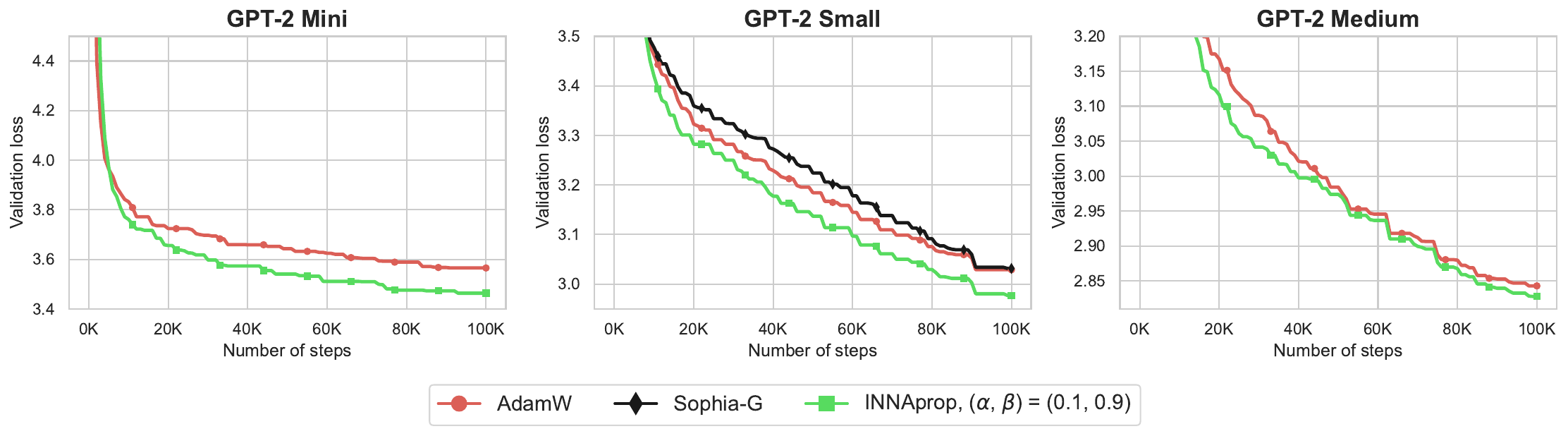}
\caption{ \footnotesize GPT-2 training from scratch on OpenWebText.}
\label{fig:gpt2-openwebtext}
\end{figure}

\paragraph{Training GPT-2 from scratch:}
We train various GPT-2 transformer models from scratch \citep{radford2019language} using the nanoGPT repository\footnote{\url{https://github.com/karpathy/nanoGPT}} on the OpenWebText dataset. For all models, gradients are clipped to a norm of 1, following \cite{mishchenko2023prodigy, liu2023sophia, brown2020language}. We use AdamW with hyperparameters from the literature \citep{liu2023sophia, brown2020language}, the standard configuration for LLM pre-training. The reported RMSprop parameter $\beta_2 = 0.95$ is different from AdamW's default (0.999), the weight decay is $\lambda=0.1$ and $\gamma_0$ depending on the network size (see \cite{brown2020language, liu2023sophia}). For example, GPT-2 small works with an initial learning rate $\gamma_0 = 6 \times 10^{-4}$. For INNAprop, we keep the same values for $\lambda$ and $\gamma_0$ as AdamW, and use the RMSprop parameter $\sigma = 0.99$ (corresponding to $\beta_2$ for AdamW), which provides the best results among values $\{0.9, 0.95, 0.99\}$ on GPT-2 mini. We use this setting for all our GPT-2 experiments with $(\alpha,\beta)=(0.1,0.9)$. The results are in \Cref{fig:gpt2-openwebtext}. INNAprop leads to a faster decrease in validation loss during the early stages compared to the baseline for GPT-2 models of Mini (30M), Small (125M), and Medium (355M) sizes. Its performance could be further improved with more thorough tuning of hyperparameters ($\alpha, \beta, \sigma, \lambda$). For GPT-2 small, we also include a comparison with Sophia-G, using the hyperparameters provided in the literature \footnote{\url{https://github.com/Liuhong99/Sophia}} \citep{liu2023sophia}.

\begin{figure}[ht!]
\centering
\includegraphics[width=\textwidth]{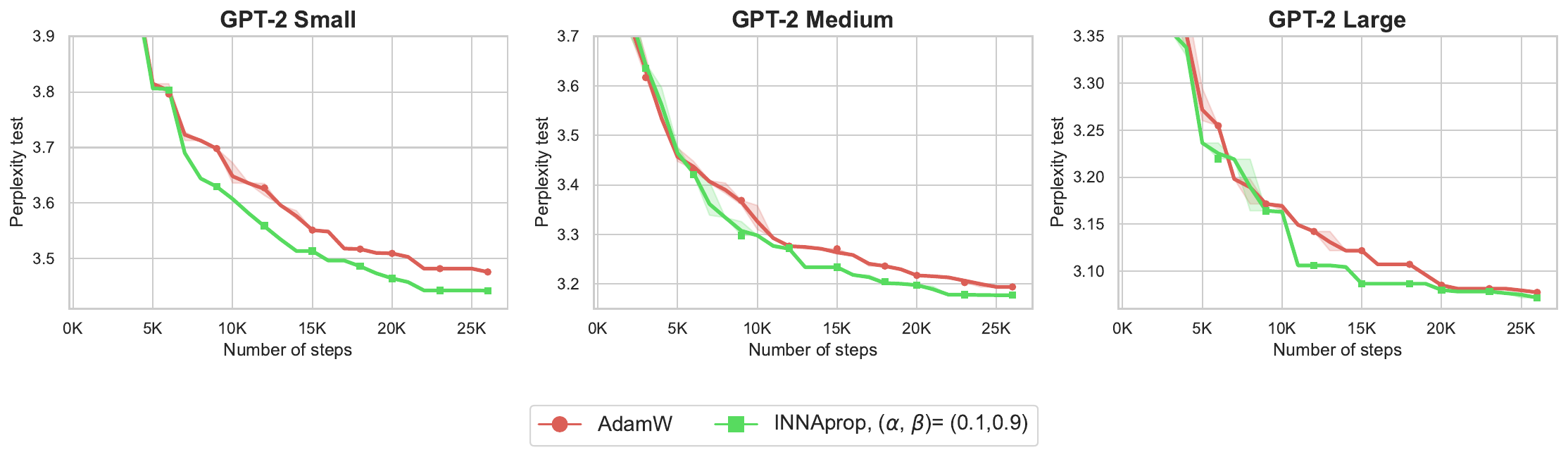}
\caption{\footnotesize Perplexity test with GPT-2 E2E Dataset with LoRA finetuning on five epochs. Three random seeds.}
\label{fig:gpt2-lora}
\end{figure}

\paragraph{Fine-tune GPT-2 with LoRA:}
Using LoRA \citep{hu2021lora}, we fine-tune the same GPT-2 models on the E2E dataset, consisting of roughly 42000 training 4600 validation, and 4600 test examples from the restauration domain. We compare AdamW and INNAprop for 5 epochs, as recommended in \cite{hu2021lora}. We use for both algorithms the same linear learning rate schedule, the recommended mini-batch size, and the RMSprop parameter ($\beta_2=\sigma=0.999$); these are listed in Table 11 in \cite{hu2021lora}. The results are displayed in Figure \ref{fig:gpt2-lora} and \Cref{tab:gpt2_table}, where we see the perplexity mean result over 3 random seeds. INNAprop with $(\alpha, \beta) = (0.1, 0.9)$ consistently achieves lower perplexity loss compared to AdamW across all GPT-2 fine-tuning experiments.

We synthetize the performance of our algorithm on LLMs below and we emphasize the capabilities of INNAprop compared to AdamW in the context of early training where gains are considerable.

\begin{table}[h!]
    \centering
    \caption{\footnotesize Performance comparison for GPT-2 training from scratch on OpenWebText (validation loss) and fine-tuning with LoRA on the E2E dataset (perplexity).}
    \label{tab:gpt2_table}
    \setlength{\tabcolsep}{4pt}
    \renewcommand{\arraystretch}{1.1}
    {\fontsize{8}{10}\selectfont{
    \begin{tabular}{l|c|c|c}
        \toprule
        \textbf{Model} & \textbf{AdamW best} & \textbf{INNAprop best} & \textbf{Steps to match AdamW} \\
        \midrule
        \multicolumn{4}{c}{\textbf{GPT-2 Training from scratch (Validation loss)}} \\
        \midrule
        GPT-2 mini & 3.57 & \textbf{3.47} & 51,000 ($1.96 \times$ faster) \\
        GPT-2 small & 3.03 & \textbf{2.98} & 79,000 ($1.26 \times$ faster) \\
        GPT-2 medium & 2.85 & \textbf{2.82} & 83,000 ($1.2 \times$ faster) \\
        \midrule
        \multicolumn{4}{c}{\textbf{GPT-2 with LoRA (Perplexity test)}} \\
        \midrule
        GPT-2 small & 3.48 & \textbf{3.44} & 19,000 ($1.31 \times$ faster) \\
        GPT-2 medium & 3.20 & \textbf{3.17} & 20,000 ($1.25 \times$ faster) \\
        GPT-2 large & 3.09 & \textbf{3.06} & 20,000 ($1.25 \times$ faster) \\
        \bottomrule
    \end{tabular}}
    }
\end{table}

\section{Conclusion}
INNAprop is an optimizer that leverages second-order geometric information while maintaining memory and computational footprints similar to AdamW. Experiments on text modeling and image classification show that INNAprop consistently matches or exceeds AdamW’s performance.

We systematically favored AdamW through the choice of recommended hyperparameters (schedulers, learning rates, weight decay). Hyperparameter tuning for friction parameters $(\alpha, \beta)$ was conducted using a grid search on CIFAR-10 (see \Cref{fig:heatmaps_resnet_wd_0.0}).  Further experiments in that direction could greatly improve the efficiency of INNAprop.

For language models, INNAprop with $(\alpha, \beta) = (0.1, 0.9)$ performs consistently well across all training durations, both for pre-training from scratch and for fine-tuning. We recommend that value for LLMs. Early training achieves notable successes (refer to \Cref{tab:gpt2_table}).

In image classification, $(\alpha, \beta) = (0.1, 0.9)$ 
 accelerates short-term learning, while higher values like $(\alpha, \beta) = (2.0, 2.0)$ improve test accuracy during longer training runs. Moreover, $(\alpha, \beta) = (2.0, 2.0)$ is effective for fine-tuning, offering a good balance between convergence speed and final accuracy. 

These experiments illustrate consistent performances of the proposed method over a diversity of benchmarks, architecture, and model scales, making INNAprop a promising competitor for the training of large neural networks.  Future research will be focused on the design of schedulers for the hyperparameters $\alpha$ and $\beta$. 

\newpage
\bibliography{references}
\bibliographystyle{iclr2025_conference}

\newpage
\appendix
\noindent This is the appendix for "A second-order-like optimizer with adaptive gradient scaling for deep learning".

\etocdepthtag.toc{mtappendix}
\etocsettagdepth{mtsection}{none}
\etocsettagdepth{mtappendix}{section}
\tableofcontents

\section{A reminder on optimization algorithms}

Considering the problem in \Cref{eq:mainProblem} and setting $\nabla\J(\theta_k) = g_k$, we outline several well-known update rule optimizers.
\begin{table*}[!ht]
{
\caption{Update rules considered for known optimizers. $\sgd$ is due to \citep{robbins1951stochastic}, $\momentum$ to \citep{polyak1964some}, $\nesterov$ to \citep{nesterov1983method}, $\rmspropmom$ to \citep{graves2013generating}, $\adam$ to \citep{kingma2015adam}, $\nadam$ to \citep{dozat2016incorporating} and $\inna$ to \citep{castera2021inertial}.}
\label{table:updaterules}
\def\arraystretch{\optarraystretch}

\rule{\textwidth}{1pt}

\vspace{0.5\baselineskip}

\begin{tabularx}{0.5\textwidth}{X}
    \sgdheader  \\
    \cmidrule(lr){1-1}
    \sgdbody  \\
    \\
    \adamheader  \\
    \cmidrule(lr){1-1}
    \adambody  \\
    \\
    \\
    \nadamheader \\
    \cmidrule(lr){1-1}
    \nadambody \\

\end{tabularx}
\begin{tabularx}{0.5\textwidth}{X}
    \momheader\\
    \cmidrule(lr){1-1}
    \mombody\\
    \\
    \rmspropmomheader\\
    \cmidrule(lr){1-1}
    \rmspropmombody\\
    \\
    \innaheader \\
    \cmidrule(lr){1-1}
    \innabody \\
\end{tabularx}
}
\vspace{0.25\baselineskip}
\rule{\textwidth}{1pt}
\end{table*}

\section{Derivation of INNAprop from DIN}
\label{sec:detailsinnaprop}

We consider \eqref{eq:dinrmsprop_start} which was a discretization of \eqref{eq:physicalIntuitionSmooth}, namely:
\begin{align}
& v_{k+1} = \sigma_2 v_{k} + (1 - \sigma_2) g_{k}^2 \\
&\frac{\theta_{k+1} - 2\theta_k + \theta_{k-1}}{\gamma^2} + \alpha \frac{\theta_{k} - \theta_{k-1}}{\gamma} + \beta \frac{\frac{g_k}{\sqrt{v_{k+1}} + \epsilon} - \frac{g_{k-1}}{\sqrt{v_k} + \epsilon}}{\gamma} + \frac{g_{k-1}}{\sqrt{v_k} + \epsilon} = 0.
\end{align}
This gives
\begin{align*}
&\frac{1}{\gamma} \left(\left(\frac{\theta_{k+1} - \theta_k}{\gamma} + \beta \frac{g_{k}}{\sqrt{v_{k+1}} + \epsilon} \right) - \left(\frac{\theta_{k} - \theta_{k-1}}{\gamma} + \beta \frac{g_{k-1}}{\sqrt{v_k} + \epsilon} \right)\right)  
=\;&- \alpha \frac{\theta_k - \theta_{k-1}}{\gamma} - \frac{g_{k-1}}{\sqrt{v_k} + \epsilon}
\end{align*}
and thus 
\begin{align*}
&\frac{1}{\gamma} \left(\left(\frac{\theta_{k+1} - \theta_k}{\gamma} + \beta \frac{g_{k}}{\sqrt{v_{k+1}} + \epsilon} \right) - \left(\frac{\theta_{k} - \theta_{k-1}}{\gamma} + \beta \frac{g_{k-1}}{\sqrt{v_k} + \epsilon} \right)\right)  \\
=\;& \left(\frac{1}{\beta} - \alpha\right) \frac{\theta_k - \theta_{k-1}}{\gamma} - \frac{1}{\beta} \left( \frac{\theta_k - \theta_{k-1}}{\gamma} + \beta \frac{g_{k-1}}{\sqrt{v_k} + \epsilon}\right).
\end{align*}
Multiplying by $\beta$, we obtain
\begin{align*}
&\frac{1}{\gamma} \left(\left(\beta\frac{\theta_{k+1} - \theta_k}{\gamma} + \beta^2  \frac{g_{k}}{\sqrt{v_{k+1}} + \epsilon} \right) - \left(\beta\frac{\theta_{k} - \theta_{k-1}}{\gamma} + \beta^2 \frac{g_{k-1}}{\sqrt{v_k} + \epsilon} \right)\right) \\
=\;& \left(1 - \alpha\beta\right) \frac{\theta_k - \theta_{k-1}}{\gamma} - \frac{\theta_k - \theta_{k-1}}{\gamma} - \beta \frac{g_{k-1}}{\sqrt{v_k} + \epsilon} 
\end{align*}
after rearranging all terms
\begin{align*}
&\frac{1}{\gamma} \left(\left(\beta\frac{\theta_{k+1} - \theta_k}{\gamma} + \beta^2 \frac{g_{k}}{\sqrt{v_{k+1}} + \epsilon}   + (\alpha \beta - 1) \theta_k \right) - \left(\beta\frac{\theta_{k} - \theta_{k-1}}{\gamma} + \beta^2 \frac{g_{k-1}}{\sqrt{v_k} + \epsilon}  + (\alpha \beta - 1) \theta_{k-1} \right)\right)   \\
=\;& - \frac{\theta_k - \theta_{k-1}}{\gamma} - \beta \frac{g_{k-1}}{\sqrt{v_k} + \epsilon}\: .
\end{align*}
Setting $\psi_{k-1} = -\beta\frac{\theta_{k} - \theta_{k-1}}{\gamma} - \beta^2 \frac{g_{k-1}}{\sqrt{v_k} + \epsilon} - (\alpha \beta - 1) \theta_{k-1}$, we obtain the recursion
\begin{align}
\label{eq:dinrmsprop_before_slot}
    v_{k+1} &= \sigma_2 v_{k} + (1 - \sigma_2) g_{k}^2 \\
    \frac{\psi_{k} - \psi_{k-1}}{\gamma} &= -\frac{\psi_{k-1}}{\beta} - \left(\alpha  - \frac{1}{\beta}\right) \theta_{k-1}\\
    \frac{\theta_{k+1} - \theta_{k}}{\gamma} &= \frac{-1}{\beta} \psi_k - \beta \frac{g_k}{\sqrt{v_{k+1}} + \epsilon} -  \left(\alpha  - \frac{1}{\beta}\right) \theta_{k}
\end{align}
We can also rewrite the above as follows:
\begin{align*}
& v_{k+1} = \sigma_2 v_{k} + (1 - \sigma_2) g_{k}^2 \\
\psi_{k+1} &= \psi_k \left(1 - \frac{\gamma}{\beta}\right) + \gamma \left(\frac{1}{\beta} - \alpha\right) \theta_k, \\
\theta_{k+1} &= \theta_k \left(1 + \gamma \left(\frac{1}{\beta} - \alpha\right)\right) - \frac{\gamma}{\beta}\psi_k - \gamma\beta \frac{g_k}{\sqrt{v_{k+1}} + \epsilon}.
\end{align*}
We can save a memory slot by avoiding the storage of $\psi_{k}$:
\begin{align}
&\quad \psi_{k+1} = \psi_k \left(1 - \frac{\gamma}{\beta}\right) + \gamma \left(\frac{1}{\beta} - \alpha\right) \theta_k, \\\notag
\Leftrightarrow &\quad \psi_{k} = \frac{\beta}{\beta - \gamma }\left( \psi_{k+1}  -  \gamma \left(\frac{1}{\beta} - \alpha\right) \theta_k \right) = \frac{\beta}{\beta - \gamma } \psi_{k+1}   - \frac{\beta}{\beta - \gamma }\gamma \left(\frac{1}{\beta} - \alpha\right) \theta_k    \\\notag
&\quad \theta_{k+1} = \theta_k \left(1 + \gamma \left(\frac{1}{\beta} - \alpha\right)\right) - \frac{\gamma}{\beta}\psi_k - \gamma\beta \frac{g_k}{\sqrt{v_{k+1}} + \epsilon} \\\notag
&\qquad \quad= \theta_k  + \gamma \left(\frac{1}{\beta} - \alpha\right)\theta_k   - \frac{\gamma}{\beta - \gamma } \psi_{k+1} + \frac{\gamma}{\beta - \gamma }\gamma \left(\frac{1}{\beta} - \alpha\right) \theta_k- \gamma\beta \frac{g_k}{\sqrt{v_{k+1}} + \epsilon} \\\notag
&\qquad \quad= \theta_k  + \left( 1 +  \frac{\gamma}{\beta - \gamma } \right) \gamma \left(\frac{1}{\beta} - \alpha\right)\theta_k   - \frac{\gamma}{\beta - \gamma } \psi_{k+1} - \gamma\beta \frac{g_k}{\sqrt{v_{k+1}} + \epsilon} \\\notag
&\qquad \quad= \theta_k  + \left(   \frac{\beta}{\beta - \gamma } \right) \gamma \left(\frac{1}{\beta} - \alpha\right)\theta_k   - \frac{\gamma}{\beta - \gamma } \psi_{k+1} - \gamma\beta \frac{g_k}{\sqrt{v_{k+1}} + \epsilon} \\\notag
&\qquad \quad= \theta_k  + \left(   \frac{\gamma  (1 - \beta\alpha)}{\beta - \gamma } \right) \theta_k   - \frac{\gamma}{\beta - \gamma } \psi_{k+1} - \gamma\beta \frac{g_k}{\sqrt{v_{k+1}} + \epsilon} \\
&\qquad \quad= \left(  1 + \frac{\gamma  (1 - \beta\alpha)}{\beta - \gamma } \right) \theta_k   - \frac{\gamma}{\beta - \gamma } \psi_{k+1} - \gamma\beta \frac{g_k}{\sqrt{v_{k+1}} + \epsilon}
\label{nostorepsik}\end{align}
Finally, we merely need to use 3 memory slots having the underlying dimension size $p$: 
\begin{align*}
& v_{k+1} = \sigma_2 v_{k} + (1 - \sigma_2) g_{k}^2 \\
\psi_{k+1} &= \psi_k \left(1 - \frac{\gamma}{\beta}\right) + \gamma \left(\frac{1}{\beta} - \alpha\right) \theta_k, \\
\theta_{k+1} & = \left(  1 + \frac{\gamma  (1 - \beta\alpha)}{\beta - \gamma } \right) \theta_k   - \frac{\gamma}{\beta - \gamma } \psi_{k+1} - \gamma\beta \frac{g_k}{\sqrt{v_{k+1}} + \epsilon}
\end{align*}

\begin{algorithm}
\caption{$\INNAprop$}
\begin{algorithmic}[1]
\STATE \textbf{Objective function:} $\J(\theta)$ for $\theta \in \mathbb{R}^p$.
\STATE \textbf{Constant step-size:} $\gamma > 0$
\STATE \textbf{Hyper-parameters:} $\sigma \in [0,1]$, $\alpha \geq 0$, $\beta > \gamma$, $\epsilon = 10^{-8}$.
\STATE \textbf{Initialization:} $\theta_0$,  $v_0 = 0$, $\psi_0 = (1 - \alpha \beta) \theta_0$.
\FOR{$k=1$ {\bfseries to} K}
    \STATE $\bm{g}_k = \nabla\J(\bm{\theta}_k)$
    \STATE $\bm{v}_{k+1} \gets \sigma \bm{v}_{k} + (1-\sigma) \bm{g}_k^2$
    \STATE $\bm{\psi}_{k+1} \gets \left(1 - \frac{\gamma}{\beta}\right)\bm{\psi}_k + \gamma \left(\frac{1}{\beta} - \alpha\right) \bm{\theta}_k$
    \STATE $\bm{\theta}_{k+1} \gets \left( 1 +  \frac{\gamma  (1 - \alpha\beta)}{\beta - \gamma} \right) \bm{\theta}_k  - \frac{\gamma}{\beta - \gamma} \bm{\psi}_{k+1} - \gamma\beta \frac{\bm{g}_k}{\sqrt{v_{k+1}} + \epsilon}$
\ENDFOR
\RETURN $\bm{\theta}_{K+1}$
\end{algorithmic}
\label{alg:innaprop_brut}
\end{algorithm}

\subsection{Equivalence between a special case of INNAprop and Adam without momentum}
\label{sec:particularcase}

In this section, we demonstrate that INNAprop with $\alpha = 1$ and $\beta = 1$ is equivalent to Adam \citep{kingma2015adam} without momentum ($\beta_1 = 0$). To illustrate this, we analyze the update rules of both algorithms. We assume that the RMSprop parameter $\beta_2$ (for Adam) and $\sigma$ (for INNAprop) are equal. Starting with INNAprop, we initialize $\psi_0 = (1 - \alpha \beta) \theta_0$. For $\alpha = 1$ and $\beta = 1$, this simplifies to $\psi_0 = 0$. The update for $\psi$ becomes:
\begin{align*}
    \psi_{k+1} &= \left(1 - \frac{\gamma}{\beta}\right) \psi_k + \gamma \left(\frac{1}{\beta} - \alpha\right) \theta_k = (1 - \gamma) \psi_k
\end{align*}
Given that $\psi_0 = 0$, it follows that $\psi_k = 0$ for all $k$. The parameter update rule for INNAprop is:
\begin{align*}
    \theta_{k+1} &= \left(1 + \frac{\gamma (1 - \alpha \beta)}{\beta - \gamma}\right) \theta_k - \frac{\gamma}{\beta - \gamma} \psi_{k+1} - \gamma \beta \frac{g_k}{\sqrt{v_{k+1}} + \epsilon}
\end{align*}
Replacing $\alpha = 1$, $\beta = 1$, and $\psi_k = 0$, we get:
\begin{align*}
    \theta_{k+1} &= \theta_k - \gamma \frac{g_k}{\sqrt{v_{k+1}} + \epsilon}
\end{align*}
Here, $g_k$ is the gradient, and $v_{k+1}$ is the exponential moving average of the squared gradients:
\begin{align*}
    v_{k+1} &= \sigma v_k + (1 - \sigma) g_k^2
\end{align*}

The Adam optimizer uses two moving averages, $m_k$ (momentum term) and $v_k$ (squared gradients):
\begin{align*}
    m_k &= \beta_1 m_{k-1} + (1 - \beta_1) g_k \\
    v_k &= \sigma v_{k-1} + (1 - \sigma) g_k^2
\end{align*}
Setting $\beta_1 = 0$, the momentum term $m_k$ simplifies to $m_k = g_k$. The update rule becomes:
\begin{align*}
    \theta_{k+1} &= \theta_k - \gamma \frac{g_k}{\sqrt{v_k} + \epsilon}
\end{align*}
This matches the form of Adam's update rule without the momentum term, confirming that INNAprop with $\alpha = 1$ and $\beta = 1$ is equivalent to Adam with $\beta_1 = 0$.

\begin{algorithm}
\caption{$\text{INNAprop with $(\alpha, \beta) =(1, 1)$}$}
\begin{algorithmic}[1]
\STATE \textbf{Objective function:} $\J(\theta)$ for $\theta \in \R^p$.
\STATE \textbf{Constant step-size:} $\gamma > 0$
\STATE{\textbf{Hyper-parameters:} $\sigma \in [0,1]$, $\alpha \geq 0$, $\beta > \gamma$, $\epsilon = 10^{-8}$}.
\STATE \textbf{Initialization:}  time step $k \leftarrow 0$, parameter vector $\theta_0$, $v_0 = 0$.
\REPEAT
    \STATE{$k \leftarrow k + 1$}
    \STATE $\bm{g}_k = \nabla\J(\bm{\theta}_k)$
    \STATE $\bm{v}_{k+1} \gets \sigma \bm{v}_{k} + (1-\sigma) \bm{g}_k^2$ 
    \STATE $\bm{\hat{v}}_{k+1} \gets \bm{v}_{k+1}/(1 - \sigma^k)$
    \STATE $\bm{\theta}_{k+1} \gets \bm{\theta}_k  - \gamma_k \left (\bm{g}_k / (\sqrt{\bm{\hat{v}}_{k+1}} + \epsilon \right)$

\UNTIL{\textit{stopping criterion is met}}
\RETURN{optimized parameters $\bm{\theta}_{k+1}$}
\end{algorithmic}
\label{alg:innaprop_alpha1_beta1}
\end{algorithm}

\begin{figure}[!ht]
    \centering
    \includegraphics[width=\textwidth]{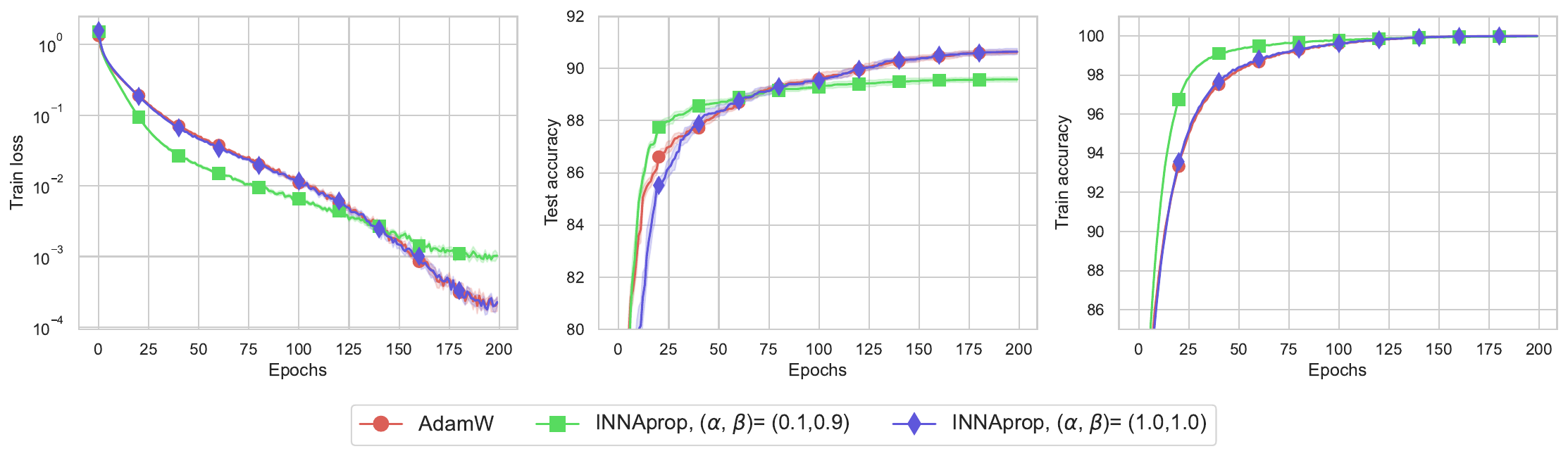}
    \caption{Training VGG11 on CIFAR10. Left: train loss, middle: test accuracy (\%), right: train accuracy (\%), with 8 random seeds.}
    \label{fig:vgg11_cifar10_alternatives}
\end{figure}

\begin{figure}[!ht]
    \centering
    \includegraphics[width=\textwidth]{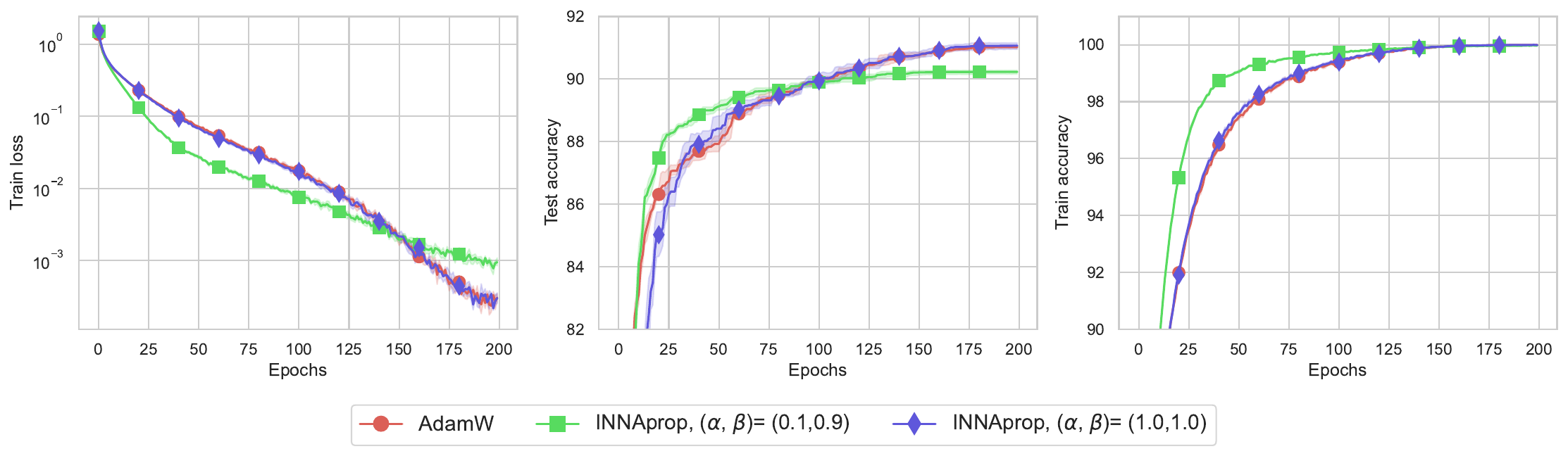}
    \caption{Training ResNet18 on CIFAR10. Left: train loss, middle: test accuracy (\%), right: train accuracy (\%), with 8 random seeds.}
    \label{fig:resnet18_cifar10_alternatives}
\end{figure}

\begin{figure}[!ht]
    \centering
    \includegraphics[width=\textwidth]{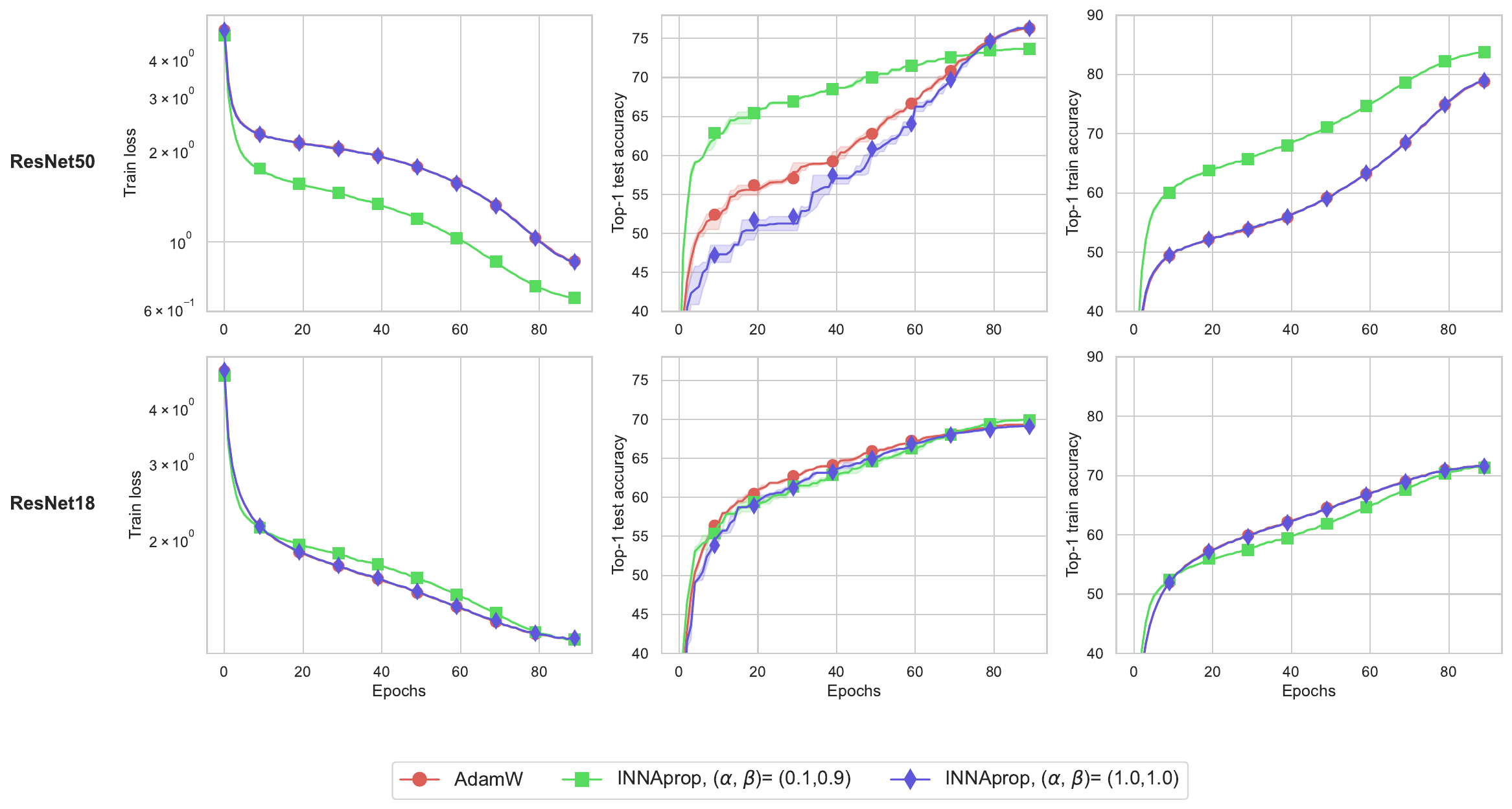}
    \caption{Training a ResNet50 (top) and ResNet18 (bottom) on ImageNet. Left: train loss, middle: Top-1 test accuracy (\%), right: Top-1 train accuracy (\%). 3 random seeds.}
    \label{fig:ResNets_alternatives}
\end{figure}

\section{Alternative discretizations} 

\subsection{An alternative derivation of INNAprop}
\label{sec:alternative_innaprop}
As mentioned in \Cref{remark:discret}, we can obtain $\INNAprop$ easily from $\inna$ \citep{castera2021inertial}. The algorithm $\inna$ writes (see \Cref{table:updaterules}): 
\begin{align*}
    \psi_{k+1} &= \psi_k + \gamma_k\left(( \frac{1}{\beta} - \alpha) \theta_k  - \frac{1}{\beta} \psi_k \right) \\
    \theta_{k+1} &= \theta_k + \gamma_k \left(( \frac{1}{\beta} - \alpha) \theta_k - \frac{1}{\beta} \psi_k - \beta g_k \right)
\end{align*}

Rearranging the terms and saving a memory slot --- use  $\psi_{k+1}$ in the second equation instead of $\psi_k$, (see \Cref{nostorepsik} for details)--- yields
\begin{align*}
    \psi_{k+1} &= \psi_k \left(1 - \frac{\gamma}{\beta}\right) + \gamma \left(\frac{1}{\beta} - \alpha\right) \theta_k \\
    \theta_{k+1} &= \left(  1 + \frac{\gamma  (1 - \beta\alpha)}{\beta - \gamma } \right) \theta_k   - \frac{\gamma}{\beta - \gamma } \psi_{k+1} - \gamma\beta g_k
\end{align*}

Now, use the $\rmsprop$ proxy directly within  $\inna$. Using the usual $\rmsprop$ constants $\sigma \in [0,1]$ and $\epsilon>0$, we obtain:

\begin{align*}
    v_{k+1} &= \sigma v_{k} + (1 - \sigma) g_{k}^2 \\
    \psi_{k+1} &= \psi_k \left(1 - \frac{\gamma}{\beta}\right) + \gamma \left(\frac{1}{\beta} - \alpha\right) \theta_k \\
    \theta_{k+1} &= \left(  1 + \frac{\gamma  (1 - \beta\alpha)}{\beta - \gamma } \right) \theta_k   - \frac{\gamma}{\beta - \gamma } \psi_{k+1} - \gamma\beta \frac{g_k}{\sqrt{v_{k+1}} + \epsilon}
\end{align*}
This is $\INNAprop$ and the derivation is much more direct, although less illustrative of the geometric features.

\subsection{A variant of INNAprop with momentum}
\label{sec:innaprop_momentum}

\paragraph{The algorithm.} We follow the rationale behind the algorithm \rmsprop\, with momentum \citep{graves2013generating}. We therefore start with  \Cref{eq:dinrmsprop_start} using the \rmsprop\, proxy for the gradient: 
\begin{align*}
& v_{k+1} = \sigma v_{k} + (1 - \sigma) g_{k}^2 \\
&\frac{\theta_{k+1} - 2\theta_k + \theta_{k-1}}{\gamma} + \alpha \frac{\theta_{k} - \theta_{k-1}}{\gamma} + \beta \frac{\frac{g_k}{\sqrt{v_{k+1}} + \epsilon} - \frac{g_{k-1}}{\sqrt{v_k} + \epsilon}}{\gamma} + \frac{g_{k-1}}{\sqrt{v_k} + \epsilon} = 0.
\end{align*}

Rearranging terms, we have 
\begin{align*}
&v_{k+1} = \sigma v_{k} + (1 - \sigma) g_{k}^2 \\
& \theta_{k+1} = \theta_k + (1 - \alpha\gamma)(\theta_k - \theta_{k-1}) - \beta\gamma \left (\frac{g_k}{\sqrt{v_{k+1}} + \epsilon} - \frac{g_{k-1}}{\sqrt{v_k} + \epsilon} \right) - \gamma^2 \frac{g_{k-1}}{\sqrt{v_k} + \epsilon}
\end{align*}

Let us introduce a momentum variable $m_k = \theta_{k-1} - \theta_k$ to obtain:
\begin{align}
v_{k+1} &= \sigma v_{k} + (1 - \sigma) g_{k}^2 \\
m_{k+1} &= (1 - \alpha\gamma) m_k + \gamma^2\frac{g_{k-1}}{\sqrt{v_k} + \epsilon} + \beta\gamma \left (\frac{g_k}{\sqrt{v_{k+1}} + \epsilon} - \frac{g_{k-1}}{\sqrt{v_k} + \epsilon} \right) \label{abc}\\ 
\theta_{k+1} &= \theta_k - m_{k+1}
\label{eq:innaprop_momentum1}
\end{align}

As previously need now to optimize the dynamics in terms of storage. For this we rewrite \Cref{abc} as
    \begin{align}
        m_{k+1} = a m_k + b g_k - c{g_{k-1}}.
        \label{eq:recursionmemory}
    \end{align}
    where $a = (1 - \alpha\gamma)$, $b = \beta \gamma$ and $c = \gamma (\beta - \gamma)$.  
Writing $\tilde{m}_k = m_k - \frac{c}{a} g_{k-1}$, we have
    \begin{align*}
        \tilde{m}_{k+1} &=  m_{k+1} - \frac{c}{a} g_k \\
        &= a m_k + b g_k - c{g_{k-1}} - \frac{c}{a} g_k \\
        &= a \left(m_k - \frac{c}{a}g_{k-1} \right) + \left(b - \frac{c}{a}\right) g_k \\
         &= a \tilde{m}_k + \left(b - \frac{c}{a}\right) g_k.
    \end{align*}
    Therefore, using this identity, we may rewrite the following 
    \begin{align*}
        m_{k+1} &= a m_k + b g_k - c{g_{k-1}},\\
        \theta_{k+1} &= \theta_k - m_{k+1}
    \end{align*}
   as
    \begin{align*}
        \tilde{m}_{k+1} &= a \tilde{m}_k + \left(b - \frac{c}{a}\right) g_k,\\
        \theta_{k+1} &= \theta_k - \tilde{m}_{k+1} - \frac{c}{a} g_{k}.
    \end{align*}

Recalling that $a = (1 - \alpha\gamma)$, $b = \beta \gamma$ and $c = \gamma (\beta - \gamma)$. Finally, we get the following recursion which is an alternative way to integrate $\rmsprop$ to $\inna$:
\begin{align}
v_{k+1} &= \sigma v_{k} + (1 - \sigma) g_{k}^2 \\\label{eq:momentinna}
\tilde{m}_{k+1} &= (1 - \alpha\gamma) \tilde{m}_k + \gamma^2 \left(\frac{1 - \alpha\beta}{1 - \alpha\gamma}\right)\frac{g_{k}}{\sqrt{v_{k+1}} + \epsilon} \\
\theta_{k+1} &= \theta_k - \tilde{m}_{k+1} - \frac{\gamma (\beta - \gamma)}{1 - \alpha\gamma}\frac{g_{k}}{\sqrt{v_{k+1}} + \epsilon}
\label{eq:innaprop_momentum2}
\end{align}
but as shown below through numerical experiments, the factor $\gamma^2$ is poorly scaled for 32 bits or lower machine precision.

\begin{figure}[!ht]
    \centering
    \includegraphics[scale=0.4]{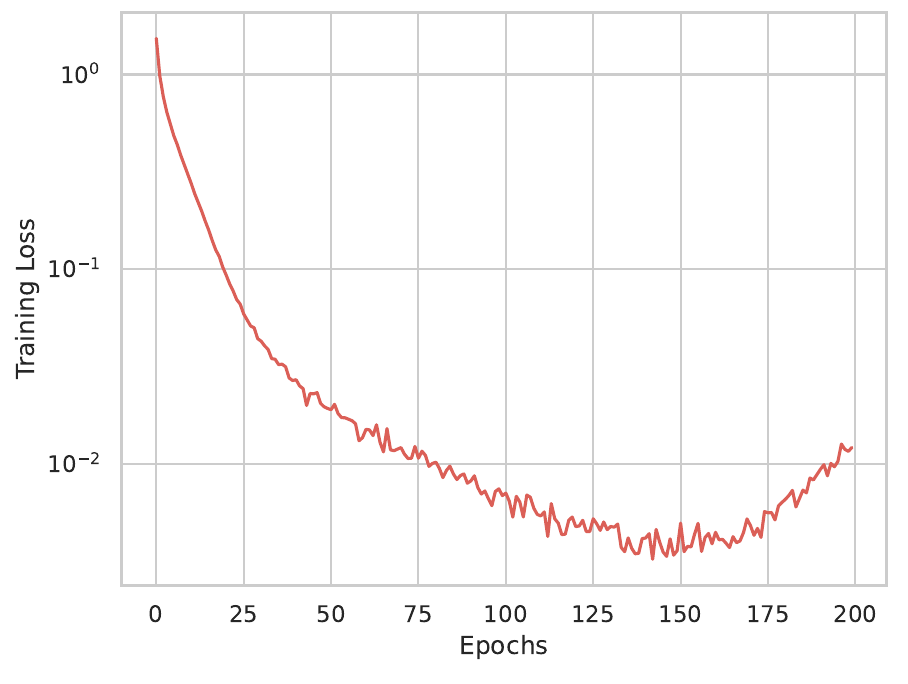}
    \caption{The version of $\inna$ with momentum of Section \ref{sec:innaprop_momentum} is an unstable method.}
    \label{fig:innaprop_momentum}
\end{figure}

\paragraph{Numerical experiments.}
Using CIFAR-10 dataset, we train a VGG11 network with the momentum version of INNAprop with the hyperparameters $(\alpha,\beta)=(0.1,0.9$) above. We used a cosine annealing scheduler with $\gamma_0=10^{-3}$ and no weight decay. As seen in Figure \ref{fig:innaprop_momentum}, the training loss stops decreasing between the 125th and 150th epochs. Upon closely examining the algorithm in this regime, we observe that at the end of training, $\gamma_{k}^2$ falls below the numerical precision, resulting in unstable behavior in \Cref{eq:momentinna}.

\subsection{An approach \`a la Adam}

\label{sec:dinadam}
In this section, we mimic the process for deriving Adam from the heavy ball with a \rmsprop\, proxy, see, e.g., \cite{kingma2015adam, ruder2016overview}, by simply replacing the heavy ball by DIN\footnote{Note that DIN with $\beta=0$ boils down to the heavy ball method.}. We call this optimizer $\dinadam$. 

%First, we discretize \eqref{eq:physicalIntuitionSmooth} according to the classical explicit Euler method. Given a solution $\theta(t)$ to Equation \eqref{eq:physicalIntuitionSmooth} at any discrete time $t_k$, we define $\theta_k = \theta(t_k)$. For the subsequent time step $t_{k+1} = t_k + \gamma_k$, where $\gamma_k$ is a small positive time step, one can approximate $\dot{\theta}_{k+1}$ and $\ddot{\theta}_{k+1}$ by 
 %\begin{align*}
%\dot{\theta}_{k+1} \simeq \frac{\theta_{k+1} - \theta_k}{\gamma_k}, \hspace{1cm}  \ddot{\theta}_{k+1} \simeq \frac{\dot{\theta}_{k+1} - \dot{\theta}_k}{\gamma_k} \simeq \frac{\theta_{k+1} - 2\theta_k + \theta_{k-1}}{\gamma_k^2}
 %\end{align*}

From \eqref{eq:physicalIntuitionSmooth}, we infer the discretization: 
\begin{equation}
\frac{\theta_{k+1} - 2\theta_k + \theta_{k-1}}{\gamma^2} + \alpha \frac{\theta_{k+1} - \theta_k}{\gamma} + \beta \frac{g_k - g_{k-1}}{\gamma} + g_k = 0.
\end{equation}

Rearranging terms, we have
\begin{equation}
\label{eq:din}
\theta_{k+1} = \theta_{k} - \frac{\gamma^2}{1 + \alpha \gamma} g_k + \frac{1}{1 + \alpha \gamma}(\theta_{k} - \theta_{k-1}) - \frac{\beta \gamma}{(1 + \alpha\gamma)}(g_k - g_{k-1})
\end{equation}
By introducing the new variable $m_k = (\theta_{k-1} - \theta_k)/\eta$ and setting $\eta>0$, we can  rewrite equation \eqref{eq:din} as:
\begin{align}
 m_{k+1} &=  \frac{1}{(1 + \alpha \gamma)}m_k + \frac{\gamma^2}{(1 + \alpha \gamma) \eta} g_k + \frac{\beta \gamma}{(1 + \alpha\gamma) \eta}(g_k - g_{k-1}) \\
\theta_{k+1} &= \theta_{k} - \eta m_{k+1}  
\end{align}

%To have an "ADAM-like" presentation, which gives a combination of $m_k$, $g_{k}$ and $g_{k-1}$, we need 

%\begin{align*}
    %\frac{1}{1 + \alpha \gamma} +  \frac{\gamma^2}{(1 + \alpha \gamma) \gamma} + \frac{\beta \gamma}{(1 + \alpha\gamma) \gamma} - \frac{\beta \gamma}{(1 + \alpha\gamma) \gamma} = 1 \\
%\end{align*}

%\jer{Je ne crois pas être d'accord avec ce qui suit, je ne vois aucune raison de lier les hyperparametres de facon à ressembler à ADAM (à moins que quelque chose m'échappe) ! Au contraire on veut que nos algo s'émancipe de l'existant. Cela voudrait dire que ce qui suit n'est pas valable. A discuter} 
To follow the $\adam$ spirit, we set $\sigma_1 = \frac{1}{(1 + \alpha \gamma) } $ and $(1- \sigma_1) =  \frac{\gamma^2}{(1 + \alpha \gamma) \eta}$. Solving for $\gamma$, we get 
\begin{align*}
    \frac{\alpha \gamma}{1 + \alpha \gamma} = \frac{\gamma^2}{(1 + \alpha \gamma) \eta} \Rightarrow \gamma = \frac{\eta}{\alpha}
\end{align*}
Then, we find the following recursion:
\begin{align}
\label{eq:dinadam}
 m_{k+1} &=  \sigma_1 m_k + (1 - \sigma_1) g_k + \beta \alpha \sigma_1 (g_k - g_{k-1})  \\
\theta_{k+1} &= \theta_{k} - \eta m_{k+1}   
\end{align}

From Equation \eqref{eq:dinadam}, we make a change of variable $\tilde{m_k} = m_k - \alpha\beta g_{k-1}$ to save a memory cell. 

\begin{align}
\label{eq:dinadam2}
\tilde{m}_{k+1} &=  \sigma_1 \tilde{m}_k + (1 - \sigma_1 + \beta\alpha\sigma_1 - \beta\alpha) g_k \\
\theta_{k+1} &= \theta_{k} - \eta (\tilde{m}_{k+1}  - \alpha\beta  g_k) 
\end{align}

Using the usual $\rmsprop$ constants $\sigma_2 \in [0,1]$ and $\epsilon>0$, we obtain:

\begin{align}
\label{eq:dinadam3}
v_{k+1} &= \sigma_2 v_{k} + (1-\sigma_2) g_k^2 \\
\tilde{m}_{k+1} &=  \sigma_1 \tilde{m}_k + (1 - \sigma_1 + \beta\alpha\sigma_1 - \beta\alpha) g_k \\
\theta_{k+1} &= \theta_{k} - \eta \frac{\tilde{m}_{k+1} - \alpha\beta g_k}{\sqrt{v_{k+1}} + \epsilon} 
\end{align}

%\jer{Il faut présenter les algo de la meme façon à chaque fois !}

\begin{algorithm}
\caption{$\dinadam$}
\begin{algorithmic}[1]
\STATE \textbf{Objective function:} $\J(\theta)$ for $\theta \in \R^p$.
\STATE \textbf{Constant step-size:} $\gamma > 0$
\STATE{\textbf{Hyper-parameters:} $(\sigma_1, \sigma_2) \in [0,1]^2$, $\alpha,\beta > 0$, $\epsilon = 10^{-8}$}.
\STATE \textbf{Initialization:} $\theta_0$,  $v_0 = 0$, $\tilde{m}_0 = 0$.
\REPEAT
    \STATE $\bm{g}_k = \nabla\J(\bm{\theta}_k)$
    \STATE $\bm{v}_{k+1} \gets \sigma_2 \bm{v}_{k} + (1-\sigma_2) \bm{g}_k^2$
    \STATE $\bm{\tilde{m}}_{k+1} \gets \sigma_1 \bm{\tilde{m}}_k + (1 - \sigma_1 + \beta\alpha\sigma_1 - \beta\alpha) \bm{g}_k$
    \STATE $\bm{\theta}_{k+1} \gets \bm{\theta}_k - \gamma \frac{\bm{\tilde{m}}_{k+1} - \alpha\beta\bm{g}_k}{\sqrt{\bm{v}_{k+1}} + \epsilon}$
\UNTIL{\textit{stopping criterion is met}}
\RETURN{optimized parameters $\bm{\theta}_k$}
\end{algorithmic}
\label{alg:dinadam}
\end{algorithm}

\begin{remark}
{\rm The way $\rmsprop$ is added in $\INNAprop$ and $\dinadam$ is different. In $\INNAprop$, RMSprop is incorporated directly during the discretization process of Equation \eqref{eq:dinrmsprop_start} for all gradients. However, in $\dinadam$, $\rmsprop$ is added only at the last step, as shown in Equation \eqref{eq:dinadam3}, and only on the gradient in the $\theta_{k+1}$ update. This is how $\rmsprop$ was combined with heavy ball to obtain $\adam$.}

\end{remark}

\begin{remark}
{\rm After setting $\alpha=1$ and $\beta=0$, we obtain $\adam$ update rules. If $\beta \neq 0$, $\dinadam$ is very close to $\nadam$ algorithm. Hence, we did not investigate this algorithm numerically.}
\end{remark}

\section{Scheduler procedures}
\label{sec:schedulers}
\textbf{Cosine annealing \citep{loshchilov2016sgdr}.} Let $\gamma_{k}$ represent the learning rate at iteration $k$, $T_{\text{max}}$ be the maximum number of iterations (or epochs), and $\gamma_{\text{min}}$ be the minimum learning rate (default value is 0). The learning rate $\gamma_k$ at iteration $k$ is given by:
$$
\gamma_k = \gamma_{\text{min}} + \frac{1}{2}(\gamma_0 - \gamma_{\text{min}})\left(1 + \cos\left(\frac{k}{T_{\text{max}}}\pi\right)\right)
$$
This scheduler was employed in all image classification experiments except for ViT.

\textbf{Cosine annealing with linear warmup \citep{radford2018improving}.} Let $\gamma_{k}$ represent the learning rate at iteration $k$, $\gamma_{\text{min}}$ the minimum learning rate, $\gamma_0$ the initial learning rate, $T_{\text{warmup}}$ the number of iterations for the warmup phase, and $T_{\text{decay}}$ the iteration number after which the learning rate decays to $\gamma_{\text{min}}$. The learning rate is defined as follows:
$$
\gamma_{k} = 
\begin{cases} 
\gamma_0\cdot \frac{k}{T_{\text{warmup}}}, & \text{if } k < T_{\text{warmup}} \\
\gamma_{\text{min}} + \frac{1}{2}\left(\gamma_0 - \gamma_{\text{min}}\right)\left(1 + \cos\left(\pi \cdot \frac{k - T_{\text{warmup}}}{T_{\text{decay}} - T_{\text{warmup}}}\right)\right), & \text{if } T_{\text{warmup}} \leq k \leq T_{\text{decay}} \\
\gamma_{\text{min}}, & \text{if } k > T_{\text{decay}}
\end{cases}
$$
This scheduler was applied in experiments involving training GPT-2 from scratch and for ViT.

%\jer{Brancher sur de Fazio ??}

\textbf{Linear schedule with linear warmup \citep{hu2021lora}.} Let $\gamma_{k}$ represent the learning rate at iteration $k$ and $T_{\text{max}}$ be the maximum number of iterations, $T_{\text{warmup}}$ be the number of warmup steps, and $\gamma_{\text{min}}$ be the minimum learning rate after warmup (default value is typically set to the initial learning rate, $\gamma_0$). The learning rate $\gamma_k$ at iteration $k$ is given by:

$$
\gamma_k = 
\begin{cases} 
\gamma_0\cdot \frac{k}{T_{\text{warmup}}} & \text{if } k < T_{\text{warmup}}, \\
\gamma_0 \cdot \left(1 - \frac{k - T_{\text{warmup}}}{T_{\text{max}} - T_{\text{warmup}}}\right) & \text{otherwise}.
\end{cases}
$$
This scheduler was used for fine-tuning GPT-2 with LoRA.
\section{Choosing hyperparameters $\alpha$ and $\beta$ for INNAprop}
\label{sec:hyperparameterTuning}

\subsection{Comparison with AdamW}
\label{sec:tuneadamw}
For VGG and ResNet training on CIFAR10, the literature suggest using initial learning rate $\gamma_0 = 10^{-3}$ with a learning rate schedule \citep{mishchenko2023prodigy, defazio2023learning, yao2021adahessian,zhuang2020adabelief}. Our experiment fix a cosine scheduler where $T_{\text{max}} = 200$ and $\gamma_{\min} = 0$ as it achieves a strong baseline for AdamW \citep{loshchilov2016sgdr, mishchenko2023prodigy}. We set weight decay $\lambda=0.1$. Then, we tune the initial learning rate $\gamma_0$ among $\{10^{-4}, 5 \times 10^{-4}, 10^{-3}, 5 \times 10^{-3}, 10^{-2}\}$. In Figure \ref{fig:optuna_adamw}, we report the performance in
terms of training loss and test accuracy for AdamW. These results confirm the usage of $\gamma_0 = 10^{-3}$.  

\begin{figure}[h]
    \centering
    \begin{subfigure}{.5\textwidth}
        \centering
        \caption{Performance rankings with VGG11.}
        \label{tab:adam_vgg11}
        \begin{tabular}{c|cc}
            \toprule
            $\gamma_0$ & \textbf{Train loss} & \textbf{Test accuracy (\%)} \\
            \midrule
            $10^{-3}$ & 0.00041 & 91.02 \\
            $5 \times 10^{-3}$ & 0.00047 & 90.86 \\
            $5 \times 10^{-4}$ & 0.00048 & 90.79 \\
            $10^{-2}$ & 0.00057 & 90.41 \\
            $10^{-4}$ & 0.00081 & 88.49 \\
            \bottomrule
        \end{tabular}
    \end{subfigure}%
    \begin{subfigure}{.5\textwidth}
        \centering
        \caption{Performance rankings with ResNet18.}
        \label{tab:adam_resnet18}
        \begin{tabular}{c|cc}
            \toprule
            $\gamma_0$ & \textbf{Train loss} & \textbf{Test accuracy (\%)} \\
            \midrule
            $10^{-3}$ & 0.00040 & 92.1 \\
            $5 \times 10^{-3}$ & 0.00049 & 91.84 \\
            $5 \times 10^{-4}$ & 0.00094 & 92.32 \\
            $10^{-2}$ & 0.00057 & 90.41 \\
            $10^{-4}$ & 0.0018 & 87.85 \\
            \bottomrule
        \end{tabular}
    \end{subfigure}
    \caption{Comparative performance of the training loss and test accuracy according to $\gamma_0$. We trained VGG11 and ResNet18 models on CIFAR10 for 200 epochs.}
    \label{fig:optuna_adamw}
\end{figure}

\section{Additional experiments}
\label{sec:additionalExperiments}

\subsection{CIFAR10 experiments}
\begin{figure}[H]
    \centering
        \includegraphics[width=\textwidth]{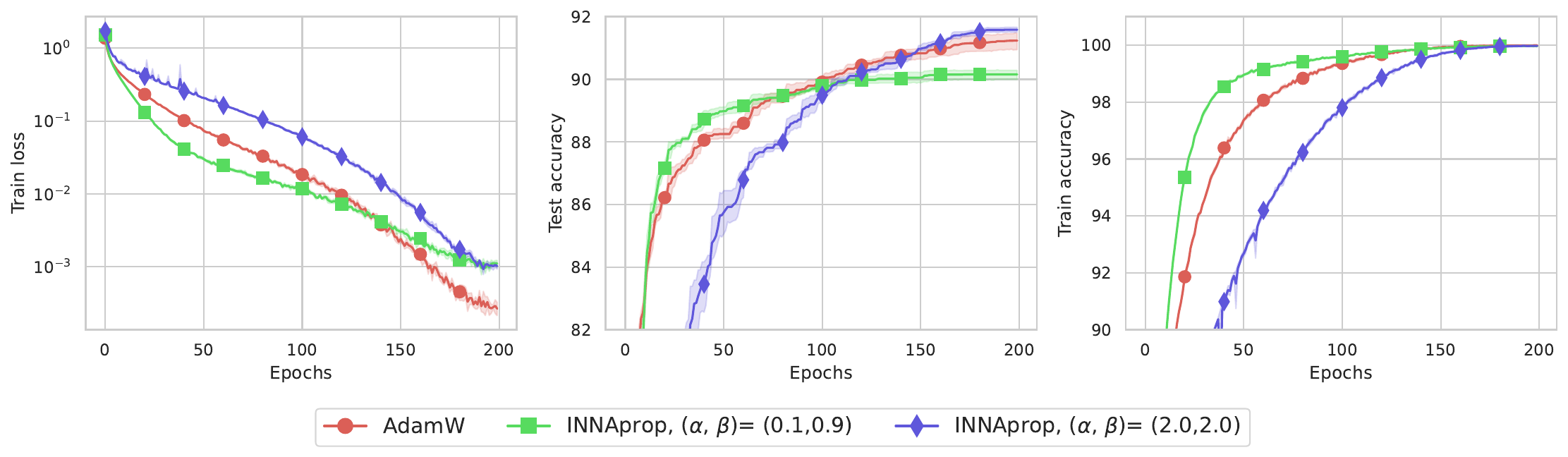}
    \caption{Training ResNet18 on CIFAR10. Left: train loss, middle: test accuracy (\%), right: train accuracy (\%), with 8 random seeds.}
    \label{fig:resnet18_cifar10}
\end{figure}

\subsection{Food101 experiments}
\begin{figure}[!ht]
    \centering
    \begin{subfigure}[b]{\textwidth}
        \centering
        \includegraphics[width=\textwidth]{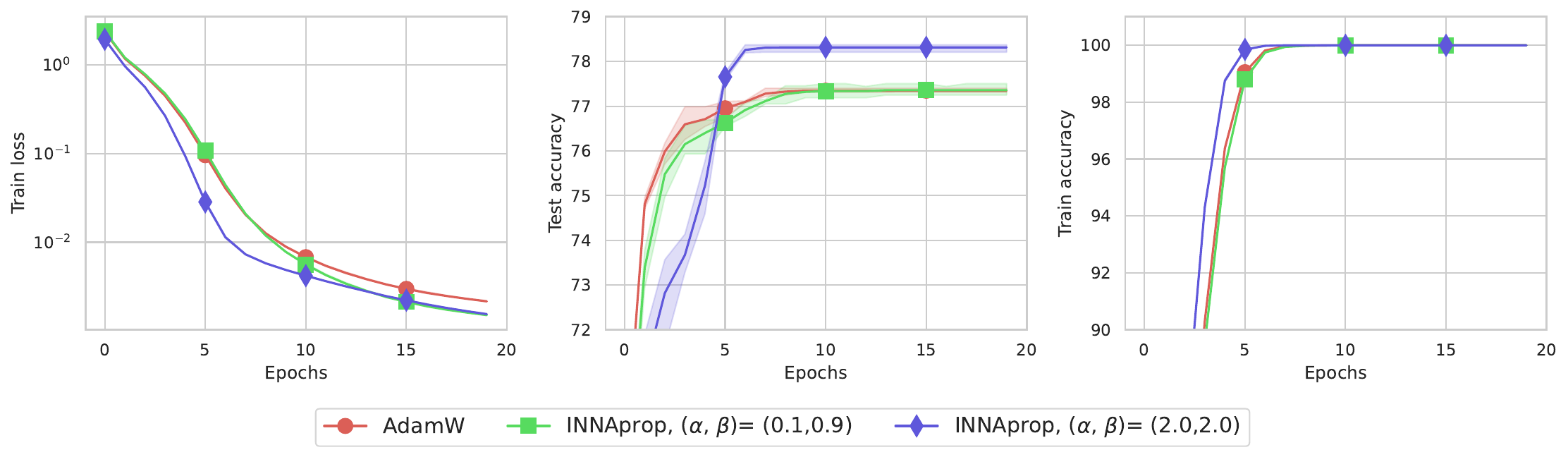}
    \end{subfigure}
    \caption{Finetuning a ResNet18 on Food101, same as \Cref{fig:food101} for ResNet18. Left: train loss, middle: test accuracy (\%), right: train accuracy (\%), with 3 random seeds.}
    \label{fig:food101appendix}
\end{figure}

\subsection{ImageNet}

\begin{figure}[!ht]
    \centering
    \includegraphics[width=\textwidth]{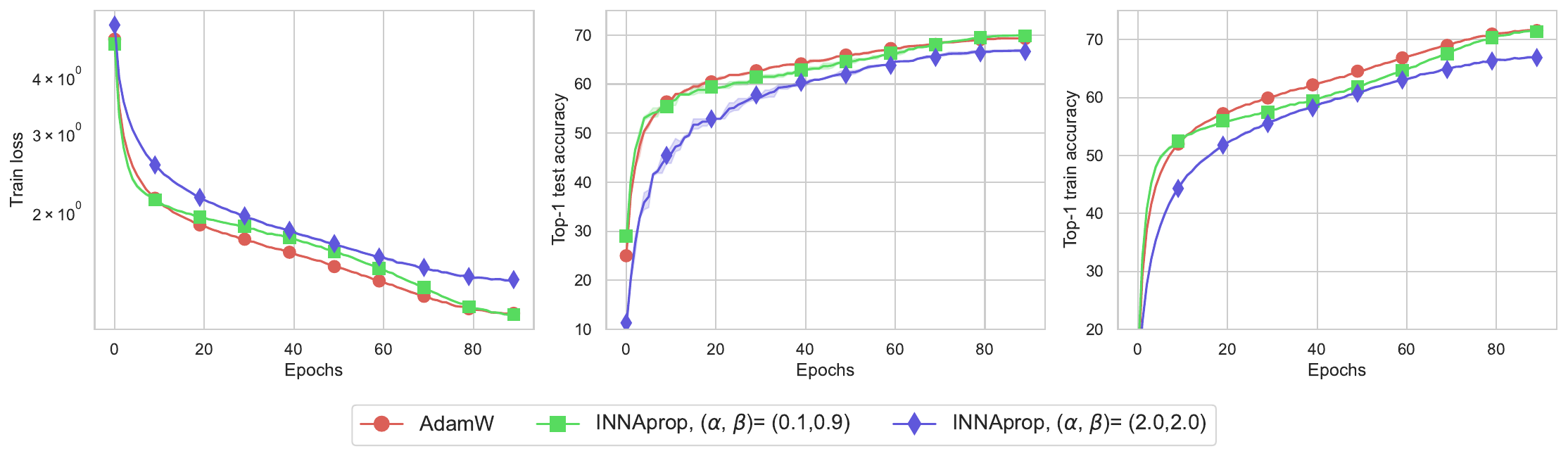}
    \caption{Training ResNet18 on ImageNet. Left: train loss, middle: test accuracy (\%), right: train accuracy (\%), with 3 random seeds.}
    \label{fig:ResNetAppendix}
\end{figure}

\begin{figure}[!ht]
    \centering
    \includegraphics[width=\textwidth]{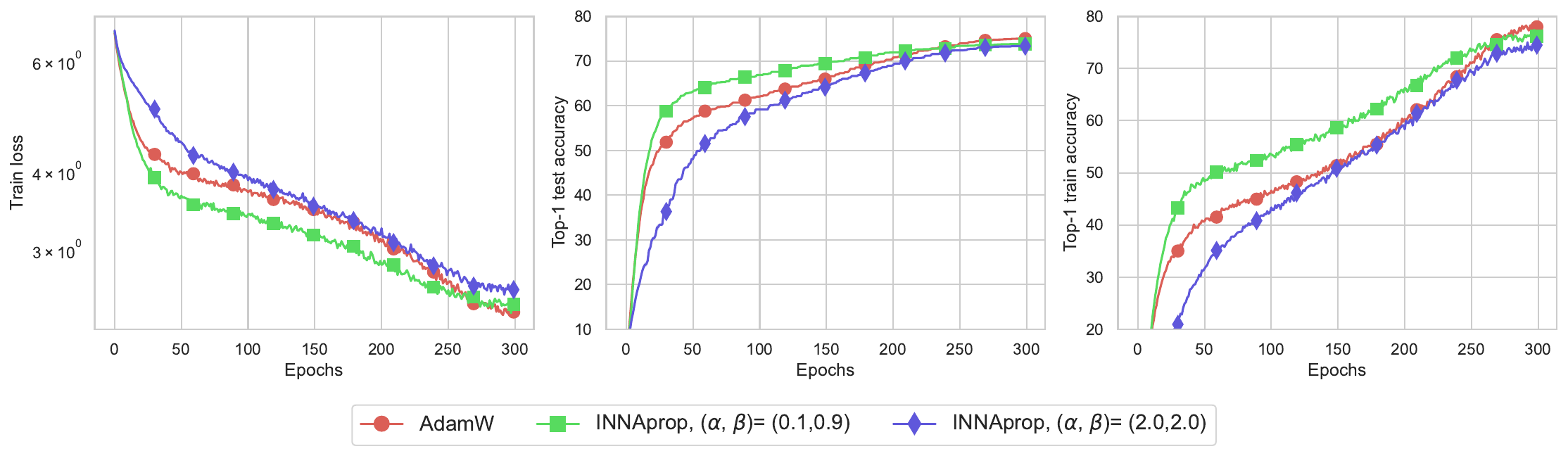}
    \caption{Fast training ViT/B-32 on ImageNet with weight decay $\lambda=0.01$ for INNAprop ($\alpha, \beta) = (0.1,0.9)$. Left: train loss, middle: test accuracy (\%), right: train accuracy (\%), with 3 random seeds.}
    \label{fig:ViTAppendix}
\end{figure}

\subsection{Heatmap for preliminary tuning of $\alpha$ and $\beta$}
\label{sec:heatmaps}
\begin{figure}[H]
    \centering
    \begin{subfigure}[b]{0.7\textwidth}
        \includegraphics[width=\linewidth]{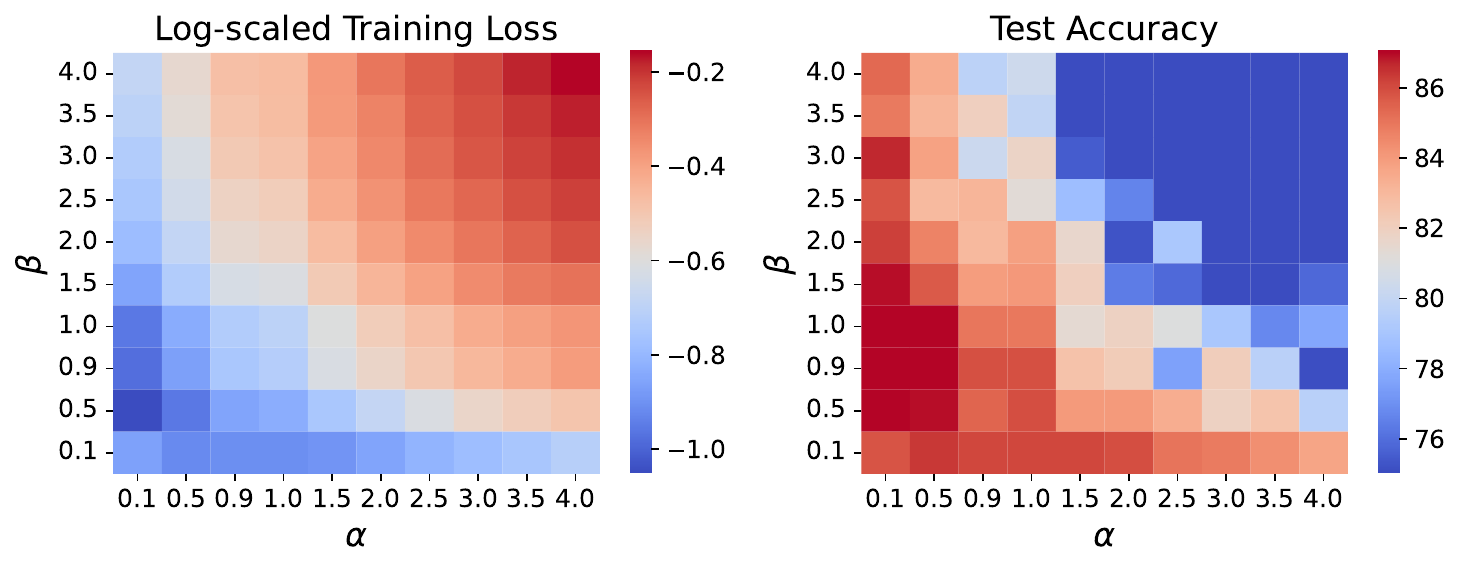}
        \caption{20 epochs}
    \end{subfigure}
    
    \begin{subfigure}[b]{0.7\textwidth}
        \includegraphics[width=\linewidth]{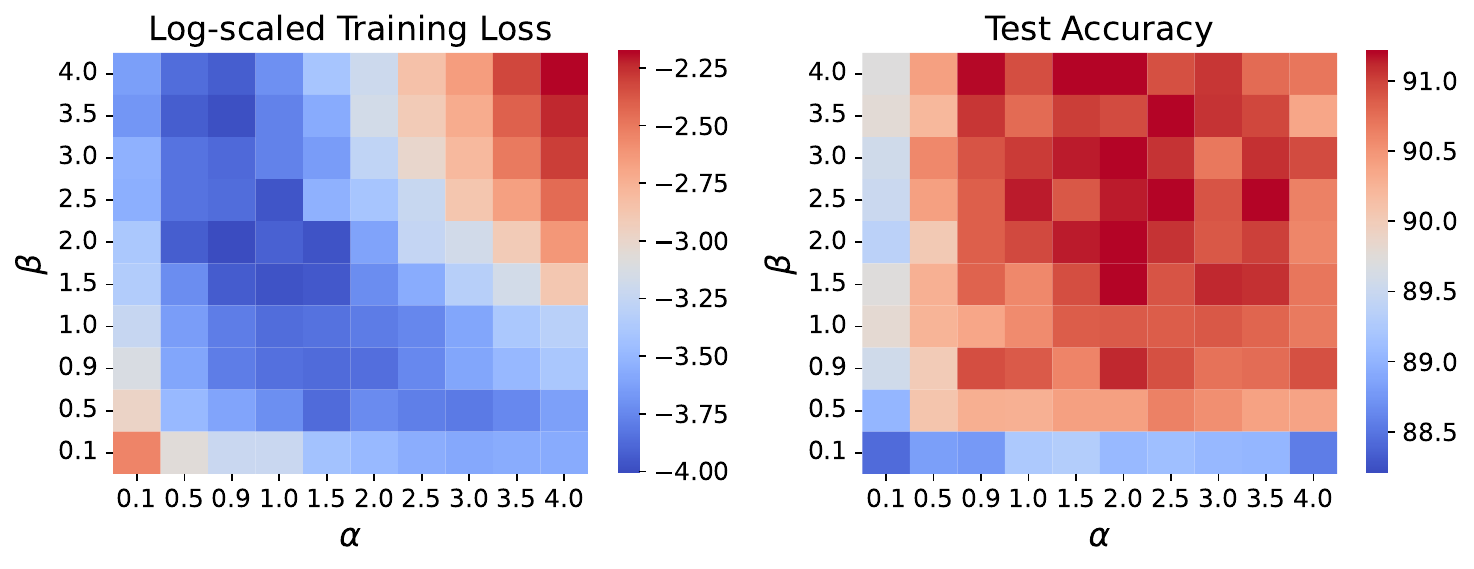}
        \caption{200 epochs}
    \end{subfigure}
    
    \caption{Log-scale training loss and test accuracies for $(\alpha, \beta)$ hyperparameters with VGG11 on CIFAR10 at different epochs. Optimal learning rate $\gamma_0 = 10^{-3}$, weight decay $\lambda = 0$.}
    \label{fig:heatmaps_vgg_wd_0.0}
\end{figure}

\begin{figure}[H]
    \centering
    \begin{subfigure}[b]{0.7\textwidth}
        \includegraphics[width=\linewidth]{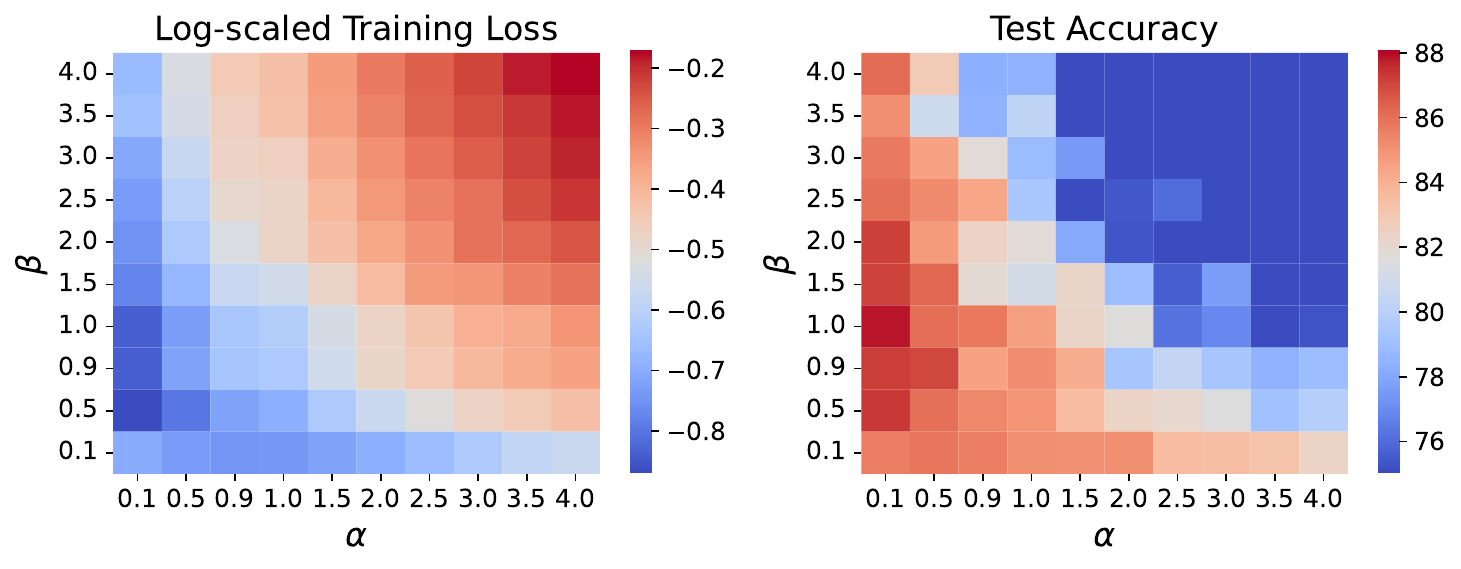}
        \caption{20 epochs}
    \end{subfigure}
    
    \begin{subfigure}[b]{0.7\textwidth}
        \includegraphics[width=\linewidth]{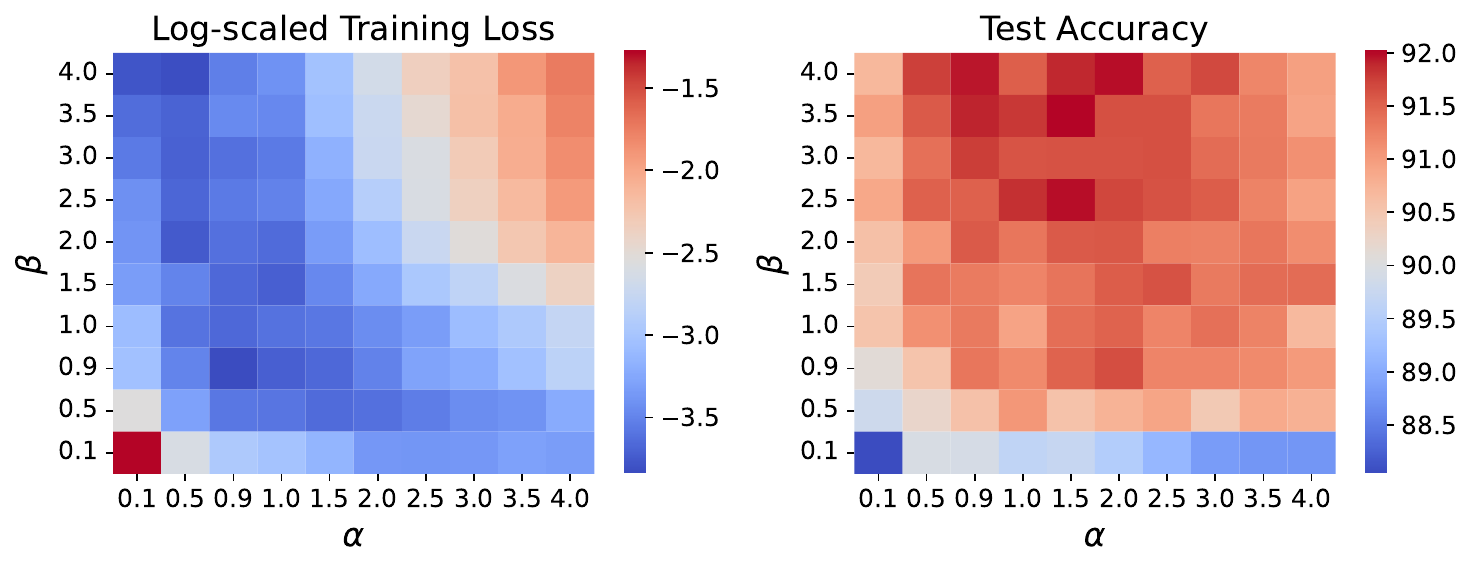}
        \caption{200 epochs}
    \end{subfigure}
    
    \caption{Log-scale training loss and test accuracies for $(\alpha, \beta)$ hyperparameters with ResNet18 on CIFAR10 at different epochs. Optimal learning rate $\gamma_0 = 10^{-3}$, weight decay $\lambda = 0.01$.}
    \label{fig:heatmaps_resnet_wd_0.01}
\end{figure}

\begin{figure}[H]
    \centering
    \begin{subfigure}[b]{0.7\textwidth}
        \includegraphics[width=\linewidth]{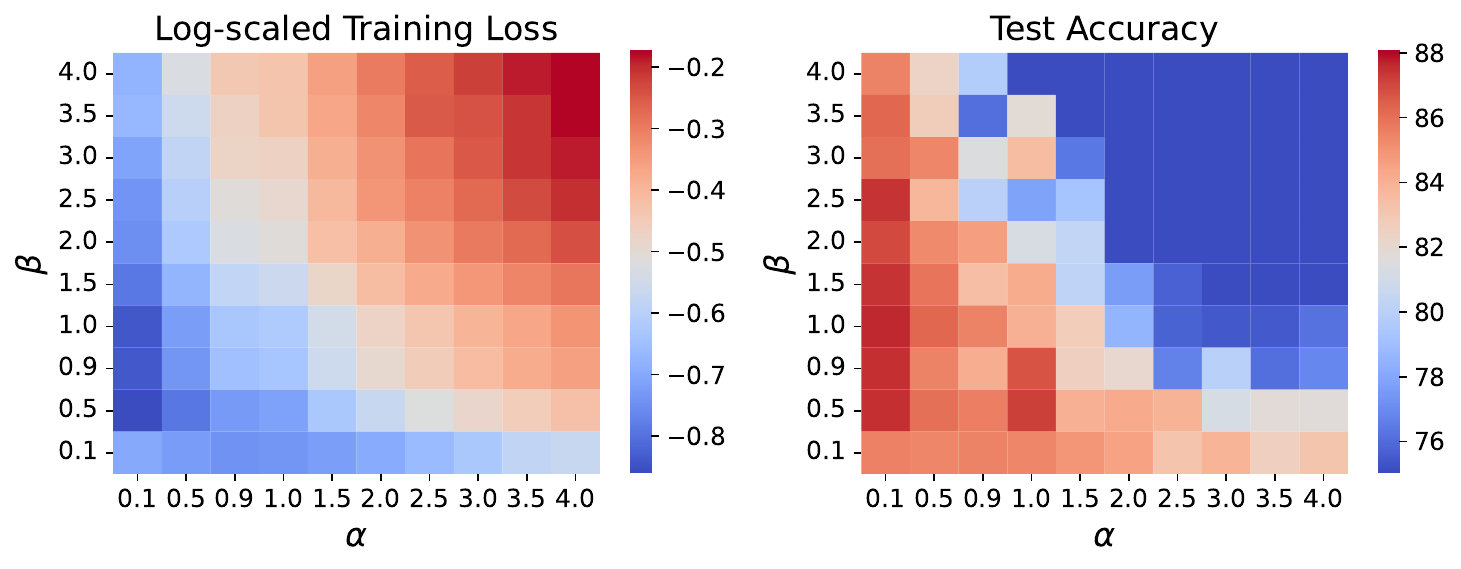}
        \caption{20 epochs}
    \end{subfigure}
    
    \begin{subfigure}[b]{0.7\textwidth}
        \includegraphics[width=\linewidth]{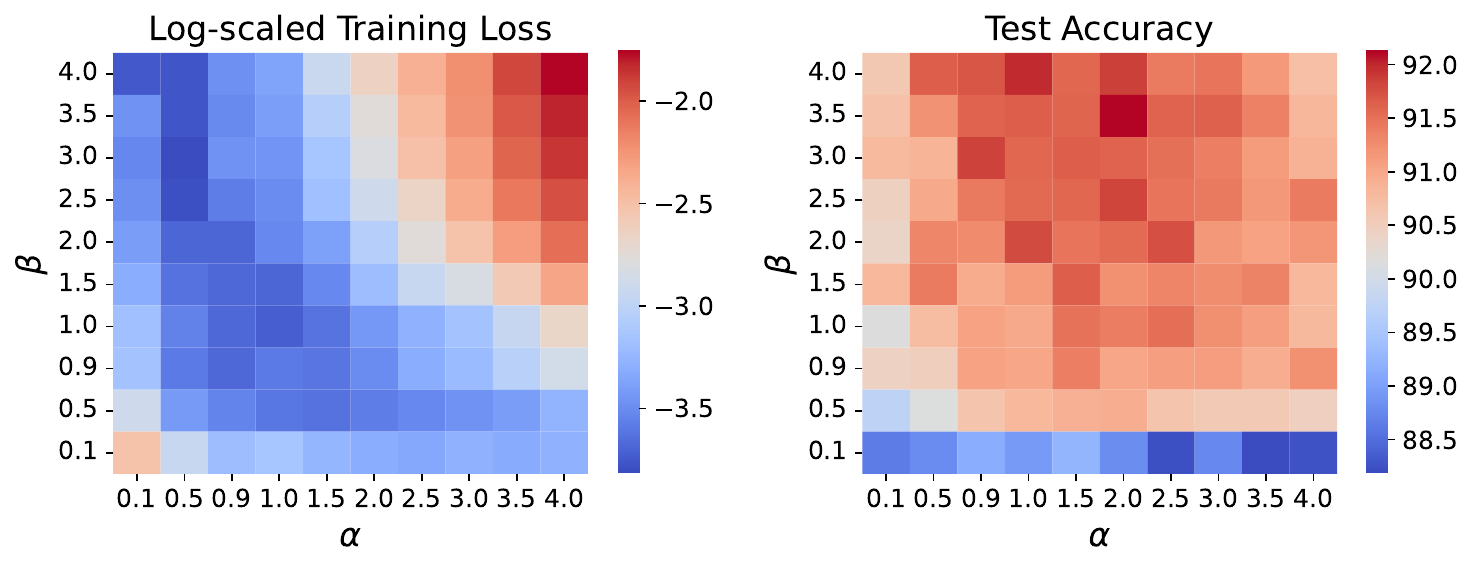}
        \caption{200 epochs}
    \end{subfigure}
    
    \caption{Log-scale training loss and test accuracies for $(\alpha, \beta)$ hyperparameters with ResNet18 on CIFAR10 at different epochs. Optimal learning rate $\gamma_0 = 10^{-3}$, weight decay $\lambda = 0$.}
    \label{fig:heatmaps_resnet_wd_0.0}
\end{figure}

\subsection{Comparision with INNA}
\begin{figure}[H]
    \centering
    \includegraphics[width=0.6\textwidth]{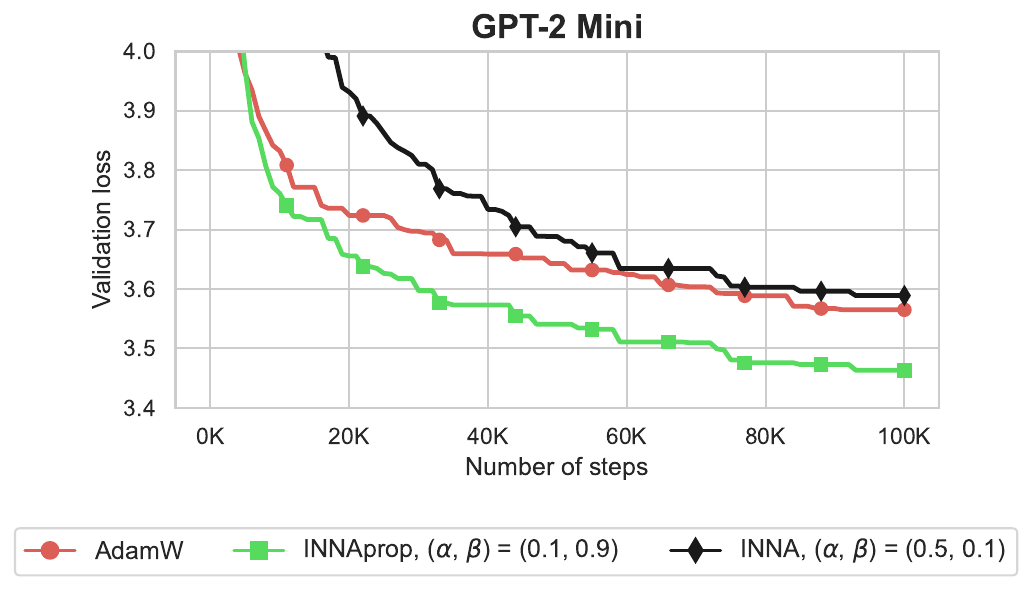}
    \caption{Validation loss comparison during GPT-2 mini training from scratch on the OpenWebText dataset.}
    \label{fig:gpt2_inna_comparison}
\end{figure}

We evaluate INNA on GPT-2 Mini and compare it to INNAprop and AdamW. Following \cite{castera2021inertial}, we used the recommended hyperparameters $(\alpha, \beta) = (0.5, 0.1)$ and tested learning rates $\{1e-4, 1e-3, 1e-2, 1e-1\}$, selecting $\gamma_0 = 0.1$ as the best. Figure \ref{fig:gpt2_inna_comparison} shows that INNAprop and AdamW outperform INNA in both convergence speed and final validation loss.

\section{Experimental Setup}
\label{sec:experiment_details}
\subsection{CIFAR-10}
We used custom training code based on the PyTorch tutorial code for this problem. Following standard data-augmentation practices, we applied random horizontal flips and random offset cropping down to 32x32, using reflection padding of 4 pixels. Input pixel data was normalized by centering around 0.5.

\begin{tabular}[t]{|c|c|}
\hline 
\textbf{Hyper-parameter}  & \textbf{Value}\tabularnewline
\hline
Architecture  & VGG11 and ResNet18\tabularnewline
\hline 
Epochs  & 200\tabularnewline
\hline 
GPUs  & 1$\times $V100\tabularnewline
\hline 
Batch size per GPU  & 256\tabularnewline
\hline  
Baseline LR  & 0.001\tabularnewline
\hline  
Seeds  & 8 runs\tabularnewline
\hline
\end{tabular}
\quad
\begin{tabular}[t]{|c|c|}
\hline 
\textbf{Hyper-parameter}  & \textbf{Value}\tabularnewline
\hline 
Baseline schedule & cosine \tabularnewline
\hline 
Weight decay $\lambda$ & 0.01\tabularnewline
\hline 
$\beta_1, \beta_2$ (for AdamW) & 0.9, 0.999\tabularnewline
\hline
$\sigma$ (for INNAprop) & 0.999 \tabularnewline
\hline
\end{tabular}

\subsection{Food101}
We used the pre-trained models available on PyTorch for VGG11 and ResNet18.\footnote{\url{https://pytorch.org/vision/stable/models.html}}. 

\begin{tabular}[t]{|c|c|}
\hline 
\textbf{Hyper-parameter}  & \textbf{Value}\tabularnewline
\hline
Architecture  & VGG11 and ResNet18\tabularnewline
\hline 
Epochs  & 200\tabularnewline
\hline 
GPUs  & 1$\times $V100\tabularnewline
\hline 
Batch size per GPU  & 256\tabularnewline
\hline  
Baseline LR  & 0.001\tabularnewline
\hline  
Seeds  & 3 runs\tabularnewline
\hline
\end{tabular}
\quad
\begin{tabular}[t]{|c|c|}
\hline 
\textbf{Hyper-parameter}  & \textbf{Value}\tabularnewline
\hline 
Baseline schedule & cosine \tabularnewline
\hline 
Weight decay $\lambda$ & 0.01\tabularnewline
\hline 
$\beta_1, \beta_2$ (for AdamW) & 0.9, 0.999\tabularnewline
\hline
$\sigma$ (for INNAprop) & 0.999 \tabularnewline
\hline
\end{tabular}
\subsection{ImageNet}
We used the same code-base as for our CIFAR-10 experiments, and applied the same preprocessing procedure. The data-augmentations consisted of PyTorch's RandomResizedCrop, cropping to 224x224 followed by random horizontal flips.
Test images used a fixed resize to 256x256 followed by a center crop to 224x224.

\subsubsection{ResNet18}

\begin{tabular}[t]{|c|c|}
\hline 
\textbf{Hyper-parameter}  & \textbf{Value}\tabularnewline
\hline
Architecture  & ResNet18\tabularnewline
\hline 
Epochs  & 90\tabularnewline
\hline 
GPUs  & 4$\times $V100\tabularnewline
\hline 
Batch size per GPU  & 64\tabularnewline
\hline 
Baseline LR & 0.001 \tabularnewline
\hline  
Seeds  & 3 runs\tabularnewline
\hline
\end{tabular}
\quad
\begin{tabular}[t]{|c|c|}
\hline
\textbf{Hyper-parameter}  & \textbf{Value}\tabularnewline
\hline 
Baseline schedule & cosine \tabularnewline
\hline  
Weight decay $\lambda$ & 0.01\tabularnewline
\hline 
$\beta_1, \beta_2$ (for AdamW) & 0.9, 0.999\tabularnewline
\hline
$\sigma$ (for INNAprop) & 0.999 \tabularnewline
\hline
\end{tabular}

\subsubsection{ResNet50}
\begin{tabular}[t]{|c|c|}
\hline 
\textbf{Hyper-parameter}  & \textbf{Value}\tabularnewline
\hline
Architecture  & ResNet18\tabularnewline
\hline 
Epochs  & 90\tabularnewline
\hline 
GPUs  & 4$\times $V100\tabularnewline
\hline 
Batch size per GPU  & 64\tabularnewline
\hline 
Baseline LR & 0.001 \tabularnewline
\hline
Mixed precision & True \tabularnewline
\hline  
Seeds  & 3 runs\tabularnewline
\hline
\end{tabular}
\quad
\begin{tabular}[t]{|c|c|}
\hline
\textbf{Hyper-parameter}  & \textbf{Value}\tabularnewline
\hline 
Baseline schedule & cosine \tabularnewline
\hline  
Weight decay $\lambda$ & 0.1\tabularnewline
\hline
$\beta_1, \beta_2$ (for AdamW) & 0.9, 0.999\tabularnewline
\hline
$\sigma$ (for INNAprop) & 0.999 \tabularnewline
\hline
\end{tabular}

\subsubsection{ViT/B-32}

\begin{tabular}[t]{|c|c|}
\hline 
\textbf{Hyper-parameter}  & \textbf{Value}\tabularnewline
\hline
Architecture  & ViT/B-32\tabularnewline
\hline 
Epochs  & 300\tabularnewline
\hline 
GPUs  & 8$\times $A100\tabularnewline
\hline 
Batch size per GPU  & 128\tabularnewline
\hline 
Baseline LR & 0.001 \tabularnewline
\hline
Seeds  & 5000 \tabularnewline
\hline
\end{tabular}
\quad
\begin{tabular}[t]{|c|c|}
\hline
\textbf{Hyper-parameter}  & \textbf{Value}\tabularnewline
\hline 
Baseline schedule & cosine \tabularnewline
\hline  
Warmup & linear for 30 epochs \tabularnewline
\hline
Weight decay $\lambda$ & 0.1\tabularnewline
\hline 
$\beta_1, \beta_2$ (for AdamW) & 0.9, 0.999\tabularnewline
\hline
$\sigma$ (for INNAprop) & 0.999 \tabularnewline
\hline
\end{tabular}

\subsection{GPT2 from scratch}
We followed the NanoGPT codebase \footnote{\url{https://github.com/karpathy/nanoGPT}} and we refer to \citep{brown2020language} as closely as possible, matching the default batch-size and schedule. 

\begin{tabular}[t]{|c|c|}
\hline 
\textbf{Hyper-parameter}  & \textbf{Value}\tabularnewline
\hline
Architecture  & GPT-2 \tabularnewline
\hline 
Batch size per gpu  &  12\tabularnewline
\hline 
Max Iters  & 100000\tabularnewline
\hline 
GPUs  & 4$\times $A100\tabularnewline 
\hline 
Dropout & 0.0\tabularnewline
\hline 
Baseline LR & refer to \citep{brown2020language}\tabularnewline
\hline 
Warmup Steps  & 500\tabularnewline
\hline 
\end{tabular}
\quad
\begin{tabular}[t]{|c|c|}
\hline 
\textbf{Hyper-parameter}  & \textbf{Value}\tabularnewline
\hline 
Seeds & 5000\tabularnewline
\hline 
Weight decay $\lambda$ & 0.1\tabularnewline
\hline
$\beta_1, \beta_2$ (for AdamW) & 0.9, 0.95 \tabularnewline
\hline 
$\sigma$ (for INNAprop) & 0.99 \tabularnewline
\hline
Gradient Clipping & 1.0 \tabularnewline
\hline
Float16 & True \tabularnewline
\hline
\end{tabular}

\subsection{GPT-2 with LoRA}
We followed the LoRA codebase \footnote{\url{https://github.com/microsoft/LoRA}} and we refer to \citep{hu2021lora} as closely as possible, matching the default batch-size, training length, and schedule. We train all of our GPT-2 models using AdamW \citep{loshchilov2017decoupled} and INNAprop on E2E dataset with a linear learning rate schedule for 5 epochs. We report the mean result over 3 random seeds; the result for each run is taken from the best epoch.

\begin{tabular}[t]{|c|c|}
\hline 
\textbf{Hyper-parameter}  & \textbf{Value}\tabularnewline
\hline
Architecture  & GPT-2 \tabularnewline
\hline 
Batch size per gpu  &  8\tabularnewline
\hline 
Epochs  & 5\tabularnewline
\hline 
GPUs  & 1$\times $A100\tabularnewline
\hline 
Dropout & 0.1\tabularnewline
\hline 
Baseline LR & 0.0002\tabularnewline
\hline 
Warmup steps  & 500\tabularnewline
\hline 
\end{tabular}
\quad
\begin{tabular}[t]{|c|c|}
\hline 
\textbf{Hyper-parameter}  & \textbf{Value}\tabularnewline
\hline 
Seeds & 3 runs\tabularnewline
\hline 
Weight decay $\lambda$ & 0.01\tabularnewline
\hline
$\beta_1, \beta_2$ (for AdamW) & 0.9, 0.98 \tabularnewline
\hline 
$\sigma$ (for INNAprop) & 0.98\tabularnewline
\hline
Learning Rate Schedule & Linear \tabularnewline
\hline
LoRA $\alpha$ & 32 \tabularnewline
\hline
\end{tabular}

\end{document}